\documentclass[elsart,12pt]{elsarticle}

\RequirePackage{amssymb}
\usepackage{graphics,epsfig}
\usepackage{amsmath,amsfonts}
\usepackage{fmtcount}
\usepackage{graphicx}
\usepackage{amsmath}
\usepackage{placeins}
\usepackage{booktabs}	
\usepackage[caption=false]{subfig}
\usepackage{algorithmic}
\usepackage{algorithm}
\usepackage{natbib}
\usepackage[usenames]{color}

\newcommand{\I}{\mathbb{I}}

\newcommand{\CRS}{\mbox{\textsc{$ret_S$}}}
\newcommand{\ACRS}{\mbox{\textsc{$RET_S$}}}
\newcommand{\ROCS}{\mbox{\textsc{$ROC_S$}}}
\newcommand{\MDDS}{\mbox{\textsc{$MDD_S$}}}
\newcommand{\SR}{\mbox{\textsc{$sharpe_S$}}}
\newcommand{\ASR}{\mbox{\textsc{$SHARPE_S$}}}
\newcommand{\RATES}{\mbox{\textsc{$RATE_S$}}}

\newcommand{\TPRMSEBETA} {\overline{TpRMSE}}

\definecolor{EditColor}{cmyk}{0.1,0.8,0,0.1}

\newtheorem{theorem}{Theorem}

\newtheorem{remark}[theorem]{Remark}

\newtheorem{definition}[theorem]{Definition}

\begin{document}
\bibliographystyle{elsarticle-num}

\newcommand{\Ret}{\mbox{\textsf{RET}}}

\newcommand{\argmin}{\operatornamewithlimits{argmin}} 
\newcommand{\cnt}[3]{\ensuremath{{#1}(\cup{#2})(\cup({#2}\cap{#3}))}}
\newcommand{\bq}{\begin{quote}}
\newcommand{\eq}{\end{quote}}
\newcommand{\be}{\begin{enumerate}}
\newcommand{\ee}{\end{enumerate}}
\newcommand{\bi}{\begin{itemize}}
\newcommand{\ei}{\end{itemize}}
\newcommand{\bd}{\begin{description}}
\newcommand{\ed}{\end{description}}

\newcommand{\mc}[1]{\ensuremath{\mathcal{{#1}}}}
\newcommand{\model}[3]{\ensuremath{{#1}=({#2},{#3})}}
\newcommand{\mon}[2]{\ensuremath{{#1}MON{#2}}}
\newcommand{\powerset}[1]{\ensuremath{\wp({#1})}}
\newcommand{\sih}[1]{\ensuremath{[ \! [ \mbox{#1} ] \! ]}}
\newcommand{\nlidb}{\textsc{Nlidb}}

\newcommand{\eqdef}{\triangleq}

\newpage

\begin{frontmatter}
	\title{Autoregressive short-term prediction of turning points using support vector regression}
	
	\author{Ran El-Yaniv}
	\ead{rani@cs.technion.ac.il}
	
	\author{Alexandra Faynburd\corref{cor1}}
	\cortext[cor1]{Corresponding author}
	\ead{alexaf@cs.technion.ac.il}

	\address{Computer Science Department, Technion - Israel Institute of Technology}
	
	\begin{abstract}
	This work is concerned with autoregressive prediction of turning points in financial price sequences. Such turning points are 
	critical local extrema points along a series, which mark the start of new swings. 
	Predicting the future time of such turning points or even their early or late
	identification slightly before or after the fact has useful applications in economics and finance.
	Building on recently proposed neural network model for turning point prediction, 
	we propose and study a new autoregressive model for predicting turning points of small swings.
	Our method relies on a known turning point indicator, a Fourier enriched representation of price histories, and support vector regression.   
	We empirically examine the performance of the proposed method
	over a long history of the Dow Jones Industrial average. Our study shows that the proposed method 
	is superior to the previous neural network model, in terms of trading performance of a simple trading
	application and also exhibits a quantifiable advantage over the buy-and-hold benchmark.
	\end{abstract}
	
\end{frontmatter}

\section {Introduction}

We focus on the difficult task of predicting turning points in financial price sequences.
Such turning points are special in the sense, that they reflect instantaneous equilibrium of demand and supply, after which 
a reversal in the intensity of these quantities takes place.
These reversals can often result from random events, in which case they cannot be predicted.
The basis for the current work is the hypothesis that numerous reversal instances are caused by
partially predictable dynamics generated by market participants.
We are not concerned here in deciphering this dynamics and extracting its mathematical laws, but merely focus on the question
of how well such turning points can be predicted in an autoregressive manner.   

Turning points can be categorized according to their ``size'', which is reflected by the duration and magnitude of the trends before and after the reversal.
Long term reversals are often termed \emph{business cycles}. A well known type of a long term reversal is 
the \emph{Kondratiev wave} (also called \emph{supercycle}), whose duration is between 40 to 60 years \cite{Kondratiev1935}. Such waves, as well as shorter term business cycles, are extensively studied in the economics literature. However, our focus here is on much shorter trend reversals whose magnitude is of order of a few percents and their cycle period is measured in days. One reason to study and predict such ``mini reversals'' is to support traders and investors in their analysis and decision making. In particular, the knowledge of future reversal 
times can help financial decision makers 
in designing safer and more effective trading strategies and can be used as a tactical aid in implementing specific trades which are motivated by other considerations.\footnote{Notice also, that the inverse of an (accurate) turning point predictor is, in fact, a trend continuation predictor.}

Our anchor to the state-of-the-art in predicting turning points is the paper by Li et al. \cite{LiDL09a} upon which we improve and expand. Hypothesizing that the underlying price formation process is governed by non linear chaotic dynamics, the Li et al.  paper proposes a model for 
short term prediction using neural networks.
They reported on prediction performance that gives rise to outstanding financial returns through a simple trading strategy 
that utilizes predictions of turning points. Li et al. casted the turning point prediction problem
as an inductive regression problem whose feature vectors consist of small windows of the most recent prices.
The regression problem was defined 
through a novel oscillator for turning points that quantifies how close in hindsight a given price is to being a peak or a trough.
Using feed-forward back propagation, they regressed the features to corresponding oscillator values and learned an ensemble of networks.
Prediction of the oscillator values was extracted from this ensemble as a weighted decision over all ensemble members.

Our contribution is two-fold. First, we replicate the Li et al. method and provide an in-depth study of their approach.
Our study invalidates some of their conclusions and confirms some others.
Unfortunately, we find that their numerical conclusions, obtained for a relatively small test set (spanning 60 days), are too optimistic.
We then consider a different learning scheme that in some sense simplifies the Li et al. approach.
Instead of ensemble of neural networks we apply Vapnik's support vector regression (SVR). 
This construction is simpler in various ways and already improves the Li et al. results by itself.
An additional improvement is achieved by considering more elaborate feature vectors,
which in addition to price data also include the Fourier coefficients (amplitude and phase) of price.
The overall new model exhibits more robust predictions that outperform the Li et al. model.
While the resulting model does exceed the buy-and-hold benchmark in terms of 
overall average return, this difference is not statistically significant.
However, the model's average Sharpe ratio is substantially better than the buy-and-hold benchmark.

\section {Related work}

Turning points have long been considered in various disciplines such as finance and economics, 
where they are mainly used for early identification of business cycles, trends and price swings.
Among the first to study turning points were Burns and Mitchell \citep{BurnsMitchell1946}, who defined a turning point in terms of business cycles with multiple stages.
Bry and Boshan \citep{Bry71} proposed a procedure for automatic detection of turning points in a time series in hindsight. 
Many of their successors refined this scheme, e.g., Pagan and Sossounov \citep{Pagan2001}.  
 
Such methods for turning points detection (in hindsight) facilitated the foundation of the study of turning points as events 
to which empirical probability can be assigned and statistical analysis can be performed. Wecker \citep{Wecker79} developed a statistical model for turning points prediction
while utilizing some of the ideas of Bry and Boshan.
Another approach was proposed by Hamilton \cite{Hamilton1989}, where he modeled turning points as switches between regimes (trends) that are governed by a two-state Markov switching model. The model was extended in \cite{Maheu2000} to include duration-dependent probabilities.

There are few works that specifically discuss turning points in stock prices. Lunde and Timmerman \citep{Lunde2004} applied a Markov Switching model with duration dependence for equities.  Bao and Yang \citep{Bao2008620} applied a probabilistic model using technical indicators as features and turning points as events of interest.  Azzini, et al. \citep{Azzini2009} used a fuzzy-evolutionary model and a neuro-evolutionary model to predict turning points. The present paper is closest to the work of Li et al. \cite{LiDL09a}, who proposed to use chaotic analysis and neural networks ensembles to forecast turning points.

\section{Preliminaries}

Let $X = x_1,x_2,\ldots,x_t,\ldots$ be a real sequence, $x_t \in \mathbb{R}$. In this paper the elements $x_t$ are economic quantities and typically 
are prices of financial instruments or indices; throughout the paper we call $x_t$ \emph{prices} and the index $t$ represents time measured in some 
time frame. Our focus is on \emph{daily} sequences
in which case $t$ is an index of a business day; our results can in principle be applied to other time frames such as weeks, hours or minutes. 
Given a price sequence $X$, we denote by $X_i^{i+N} = x_i,\ldots,x_{i+N}$
a consecutive subsequence of $N$ prices that starts at the $i$th day.

Throughout the paper we will consider autoregressive prediction mechanisms defined for price sequences $X$.
For each day $t$ we consider a recent window of $m$ prices, $W_t = X_{t-m}^{t-1}$ called the \emph{backward window} of day $t$. 
The prices in the backward window may be transformed to a \emph{feature space} of cardinality $n$ via some encoding transformation.

\subsection {Turning points and their properties}
\label{sec:ProblemSetting}
\begin{figure}[ht]
  \centering
	\includegraphics[scale=0.7]{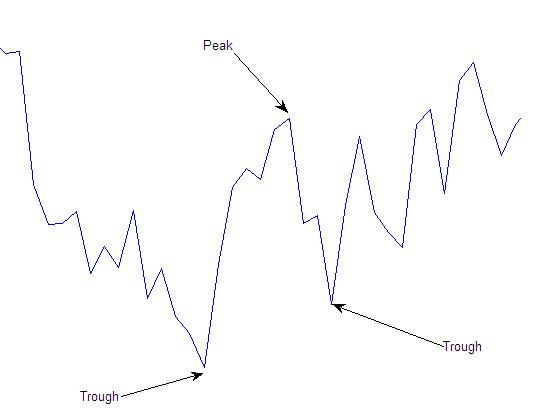}
	\caption{Examples for peaks and troughs}
	\label{fig:PeakAndThrough}
\end{figure}

Let $X$ be a price sequence.
A \emph{turning point (TP)} or a \emph{pivot} in $X$ is a time index $t$ where a local extremum (either minimum or maximum)
is obtained.
A turning point is called a \emph{peak} if it is a local maximum, and a \emph{trough} if it is a local minimum.
Examples for peaks and troughs are shown in Figure~\ref{fig:PeakAndThrough}.

Ignoring commissions and other trading ``friction,''a trader who is able to buy at troughs and sell at peaks, 
i.e., to enter/exit the market precisely at the turning points, would gain the maximum possible profit. 
For this reason, successful identification and forecasting of turning points is extremely lucrative.
However, even if all troughs and peaks were known in hindsight, 
due to friction factors  including commissions, bid/ask spreads, trading liquidity and latency, attempting to
exploit all fluctuations including the smallest ones may result in a loss. 
Therefore, one of the first obstacles when attempting to exploit turning points is to define the target 
fluctuations we are after, so as to ignore the smaller sized ones.
To this end, we now consider three definitions, each of which can quantify the ``size'' of turning points.

\begin{figure}[htbp]
  \centering
  \subfloat[Pivot point (trough) of \newline degree $\ge10$]{\label{fig:PivotTurningPointScheme}\includegraphics[width=0.3\textwidth]{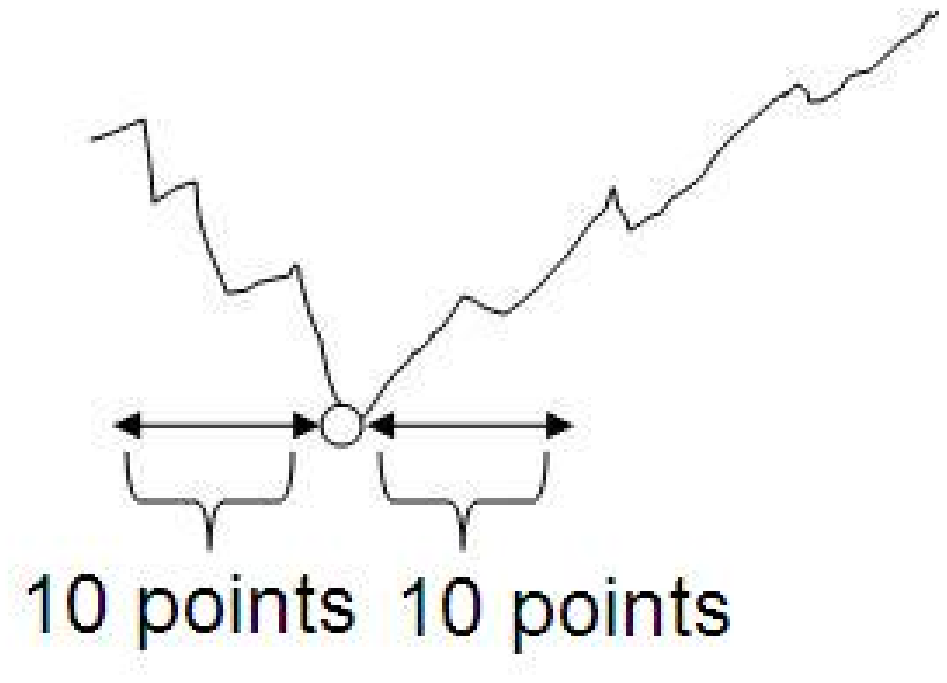}}                 \subfloat[Turning point of impact $\gamma$]{\label{fig:ImpactTurningPointScheme}\includegraphics[width=0.3\textwidth]{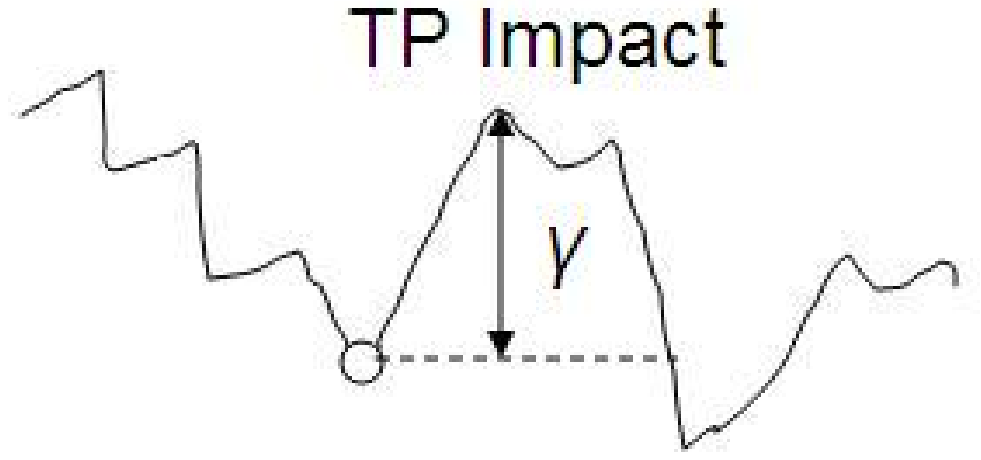}}
  \subfloat[Turning point with momentum $\gamma$ with respect to a lookahead window of length  $w$]{\label{fig:MomentumTurningPointScheme}\includegraphics[width=0.3\textwidth]{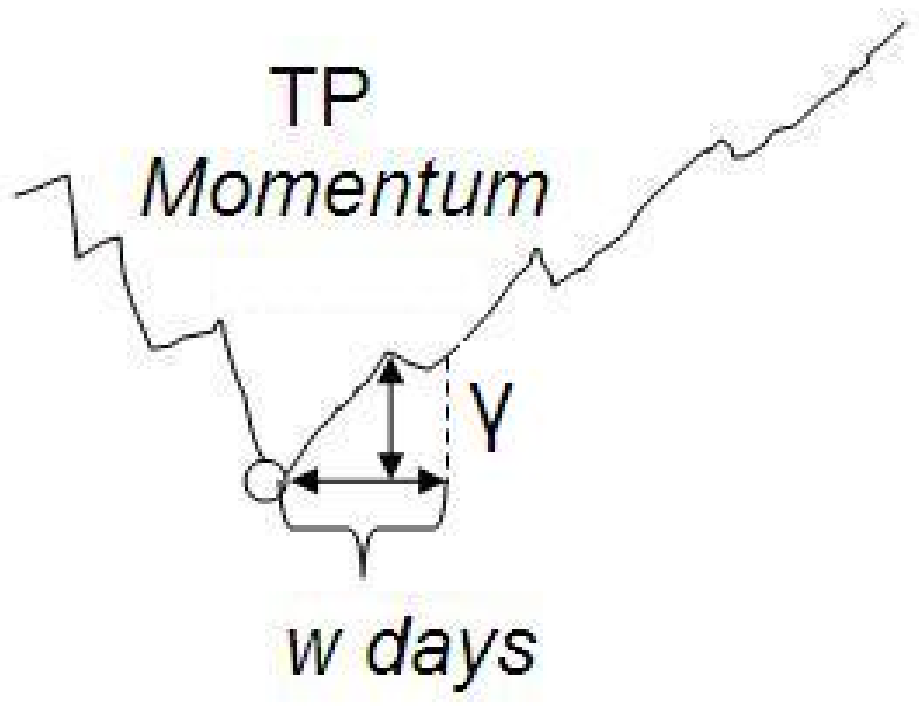}}
  \caption{Turning points schematic examples}
  \label{fig:TurningPointSchemes}
\end{figure}

\begin{definition}[Pivot of degree $K$]
The time index $t$ in a time series is an \emph{upper pivot} or a  \emph{peak} of degree $K$, if for all $j = 1,2,...,K$, $x_t > x_{t-j}$
and $x_t > x_{i+j}$. Similarly, $t$ is a \emph{lower pivot} or a \emph{trough} of degree $K$, if for $j = 1,2,...,K$, $x_t < x_{t-j}$
and $x_t < x_{t+j}$.
\end{definition}
A trough of degree 10 is depicted schematically in Figure \ref{fig:PivotTurningPointScheme}.
A \emph{turning point} is any pivot of degree at least 1. 
By definition, at time $t$ one has to know the future evolution of the sequence for the following $K$ days in order to determine if $t$ 
is a pivot of degree $K$. Typically, pivots of higher degree correspond to larger price swings. Therefore, such pivots are harder to
identify in real time.

The following two definitions consider two other properties of pivots that reflect their ``importance''.
These properties will be used in our applications. Definition~\ref{def:ImpactTurningPoint} is novel and Definition~\ref{def:SpeedTurningPoint}
is due to \citep{LiDL09a}.
 
\begin{definition}[Impact of a turning point]
	\label{def:ImpactTurningPoint}
	The \emph{upward impact} of a trough $t$ is the 
	ratio
	$
	\max\{x_t,\ldots,x_n\} / x_t,
	$
	where $n$ is the first index, greater than $t$, such that $x_n < x_t$.
	That is, if the sequence increases after the trough $t$ to some maximal value, $x_{max}$, and then decreases below $x_t$;
	the impact is the ratio $x_{max}/x_t$.
	If $x_t$ is the global minimum of the sequence, then the numerator is taken as the global maximum appearing after time $t$.
	The \emph{downward impact} of a peak is defined conversely.
	\end{definition}

\begin{definition}[Momentum of a turning point \citep{LiDL09a}]
	\label{def:SpeedTurningPoint}
	The \emph{upward momentum} of a trough $t$ with respect to a lookahead window of length $w$ is the 
percentage increase from $x_t$ to the maximal value in the window $X_{t+1}^{t+w} = x_{t+1},\ldots,x_{t+w}$. 
That is the upward momentum is $\max\{x_{t+1},\ldots,x_{t+w}\} / x_t$.
The \emph{downward momentum} of a peak is defined conversely.
	\end{definition}

\begin{figure}[htbp]
	\subfloat[Turning points  (pivots) of  degree K=10]
	{
	\includegraphics[scale=0.4]{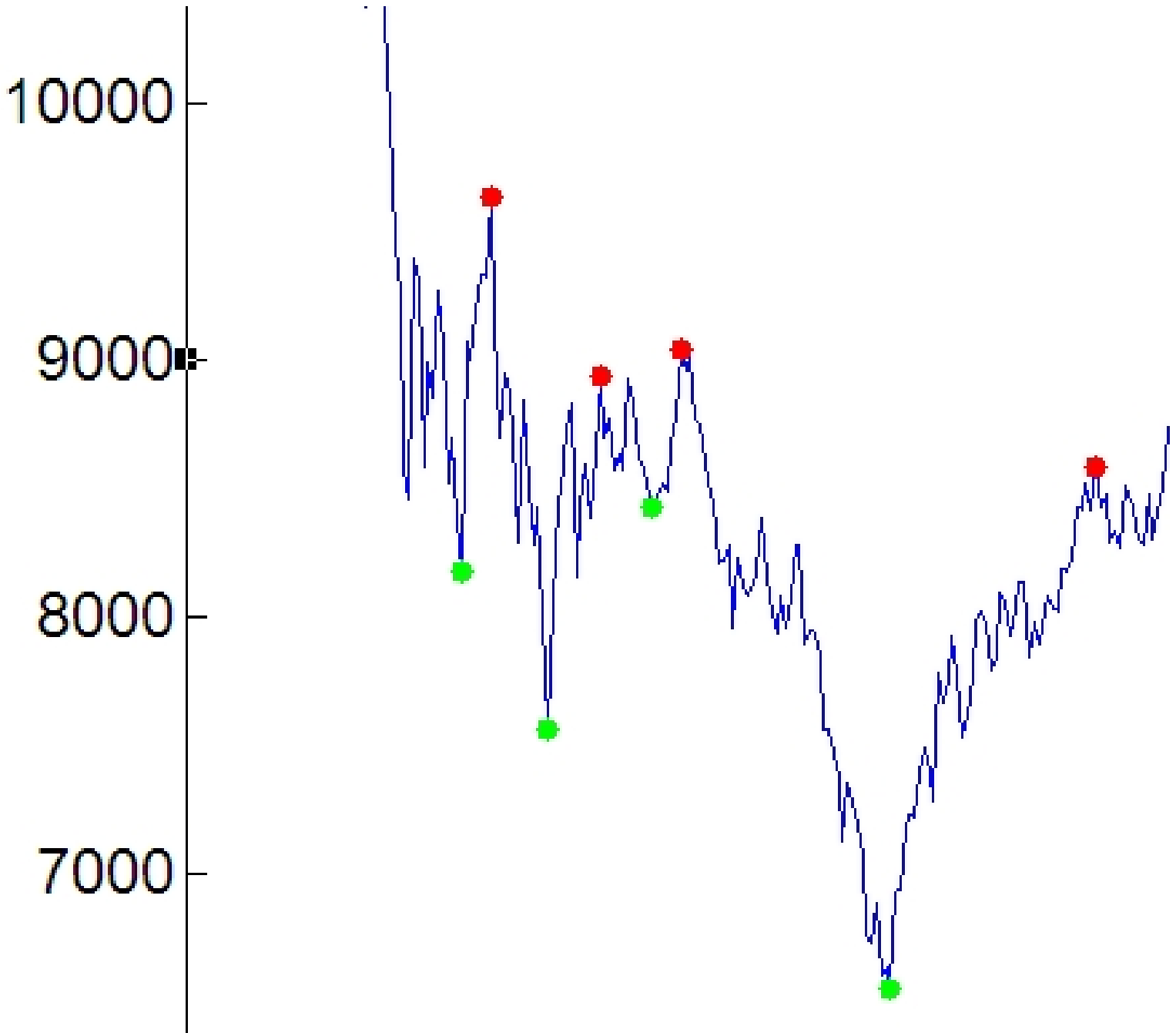}
	\label{fig:PivotDegreeTurningPointExample}
	}
	\qquad
	\subfloat[Turning points with impact $\gamma=10\%$]
	{
			\includegraphics[scale=0.4]{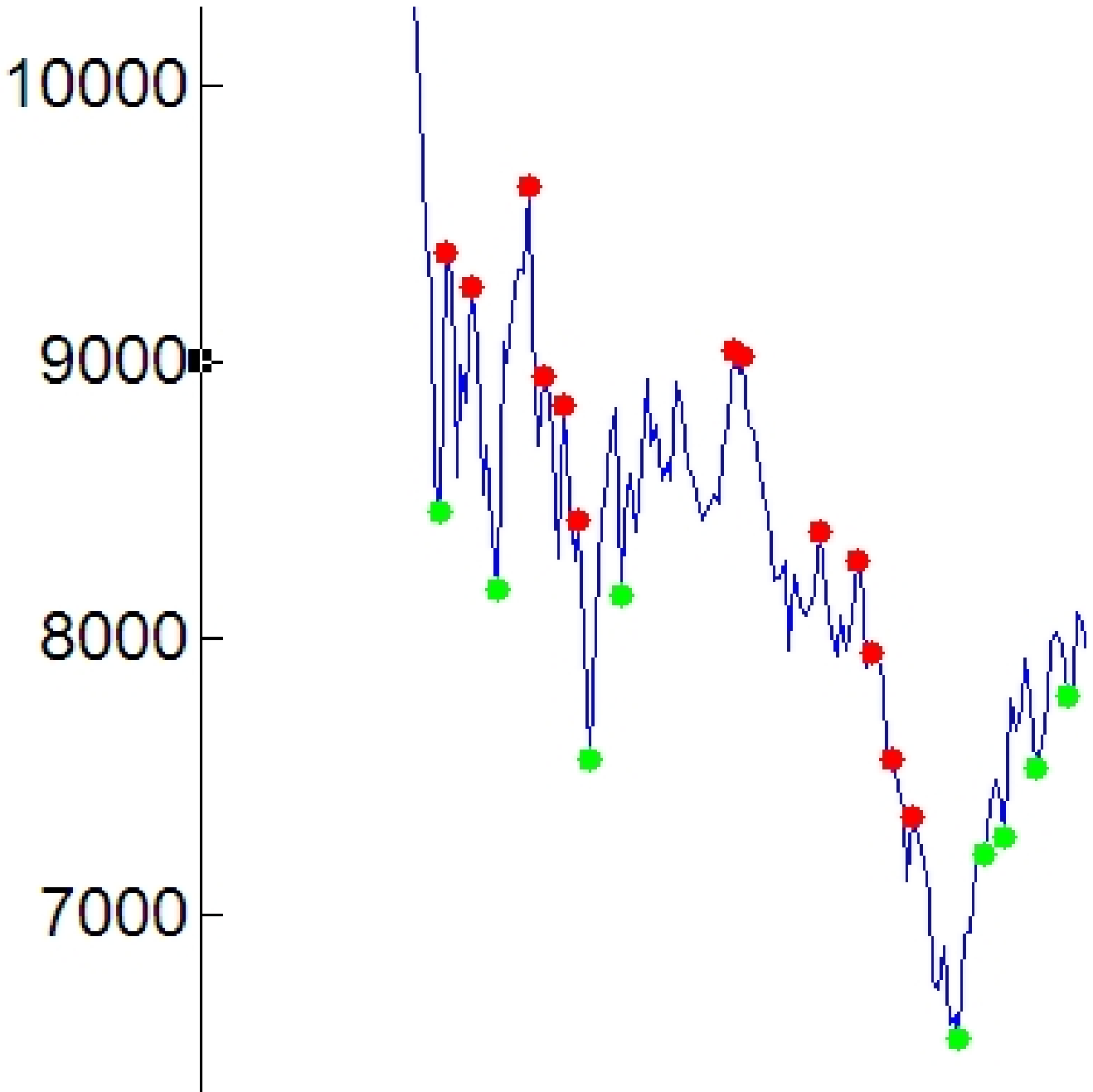}
		  \label{fig:ImpactTurningPointExample}
	}
	\qquad
	\begin{center}
	\subfloat[Turning points with momentum $\gamma$= 10\% with respect to a lookahead window of length $w$=6 days]
	{
		\includegraphics[scale=0.4]{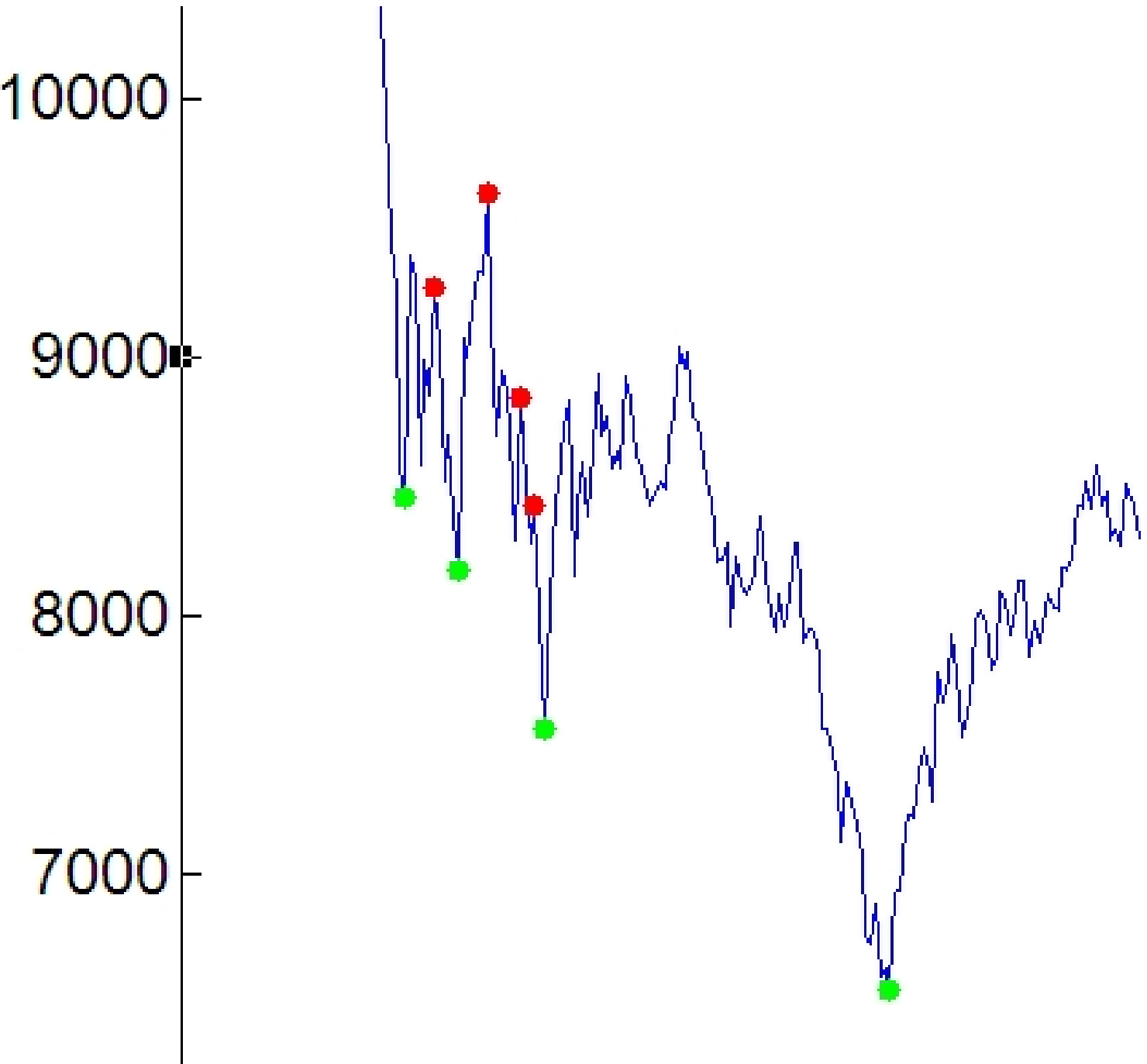}
		\label{fig:MomentumTurningPointExample}
	}
	
	\end{center}
	 \caption{Examples of turning point types over the DJIA, 5/2008-5/2009}
	 
	 \label{fig:TurningPointExamples}
\end{figure}

In Figure~\ref{fig:TurningPointSchemes} we schematically depict these three characteristics (i.e., pivot degree, impact and momentum).
Then, in Figure~\ref{fig:TurningPointExamples} we show examples on a real price sequence. Notice, that these definition give rise to 
quite different turning points identification.

\subsection{Alternating pivots sequence}
\label{sec:altPivots}

\begin{figure}[hbp]
  \begin{center}
  	\subfloat[Find all the turning points of momentum $\gamma=0.05, w =6$]
  	{
  		\label{fig:TurningPointsSelection1}
  		\includegraphics[scale=0.7]{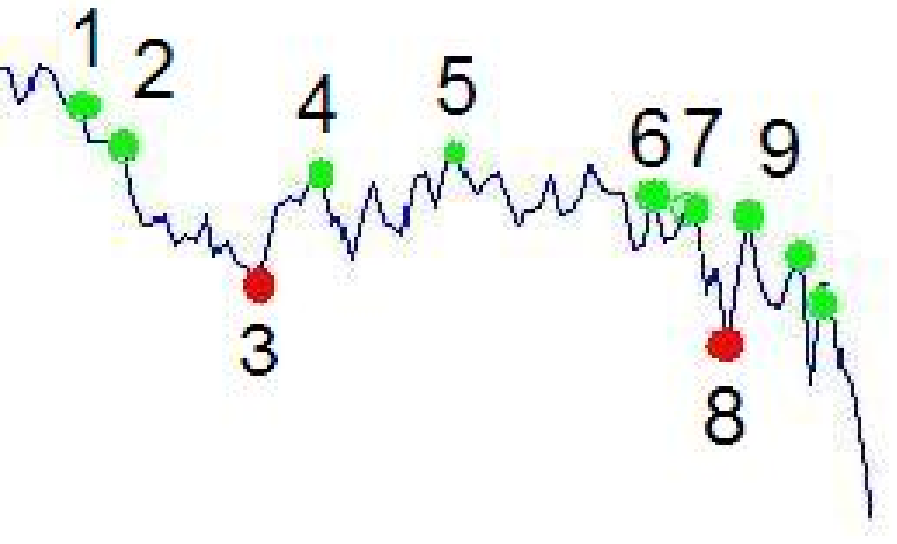}
  	}
  
  	\subfloat[Ensure alternation of peaks and troughs. Points 2,5,6,7 are eliminated]
  	{
	  	\label{fig:TurningPointsSelection2}
	  	\includegraphics[scale=0.7]{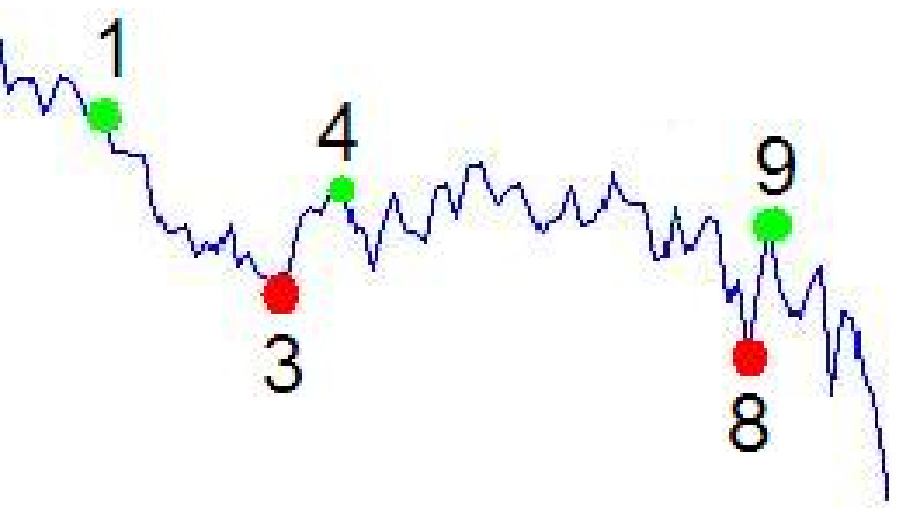}
  	}
  
  	\subfloat[A peak should represent the highest point between the troughs. Point 4 is replaced with point 5 as a maximal TP with the defined properties between the troughs 3 and 8]
  	{
  		\label{fig:TurningPointsSelection3}
  		\includegraphics[scale=0.7]{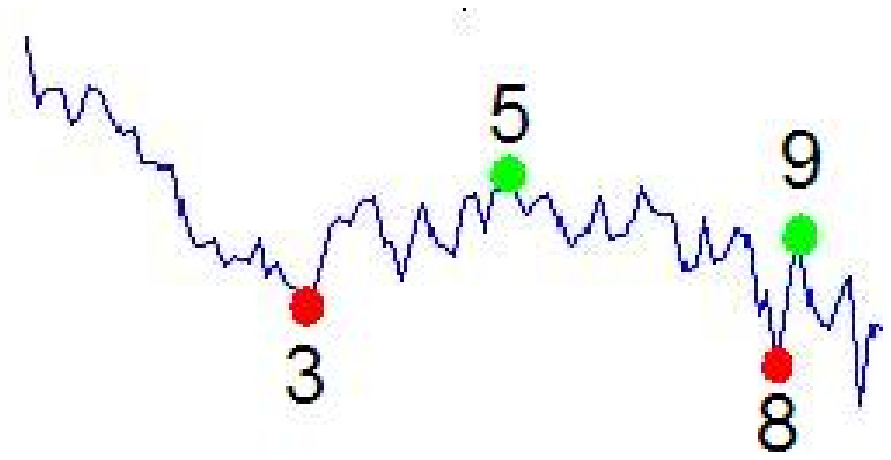}
  	}
  \end {center}
  
  \caption{Turning points selection example for momentum turning points}
  \label{fig:TurningPointsSelection}
  
\end{figure}

Given a time series $X$ and required characteristics of turning points (e.g.,  pivots of degree 10, or pivots with impact $\gamma$, etc.)
we would like to extract from $X$ an alternating sequence $A(X)$ of turning points. 
The sequence $A(X)$ is then used to construct a turning point oscillator (in Section~\ref{sec:TurningPointOscillator}). 
Following \cite{Bry71}, we require that the alternating sequence $A(X)$ 
will satisfy the following  requirements:
\begin{enumerate}
	\item Only pivots with the required characteristics will be included in $A(X)$.
	\item The pivot sequence will alternate between peaks and troughs.
	\item With the exception of first and last elements,  every trough will correspond to a global minimum in the time interval
	defined by the pair of peaks surrounding it, and vice versa --
	every peak is a global maximum in the time interval defined by the troughs surrounding it.
\end{enumerate}
In \ref{sec:TPExtractionDetails} we present an algorithm that extracts an alternating pivot sequence that satisfies the above conditions.
In Figure \ref{fig:TurningPointsSelection} an example is given, showing three steps of this algorithm 
corresponding to the above three requirements.
The proposed algorithm is by no means the only way to compute a proper alternating sequence.
We note, however, that any algorithm that extracts an alternating pivot sequence as defined above must rely on hindsight.
Therefore, in real time applications the use of the algorithm is restricted to training purposes.

\begin{figure}[!ht]
  \centering
  \subfloat[Turning points of impact $\gamma=0.03$]
  {
  	\label{fig:PotentialTurningPoints}
  	\includegraphics[width=0.5\textwidth]{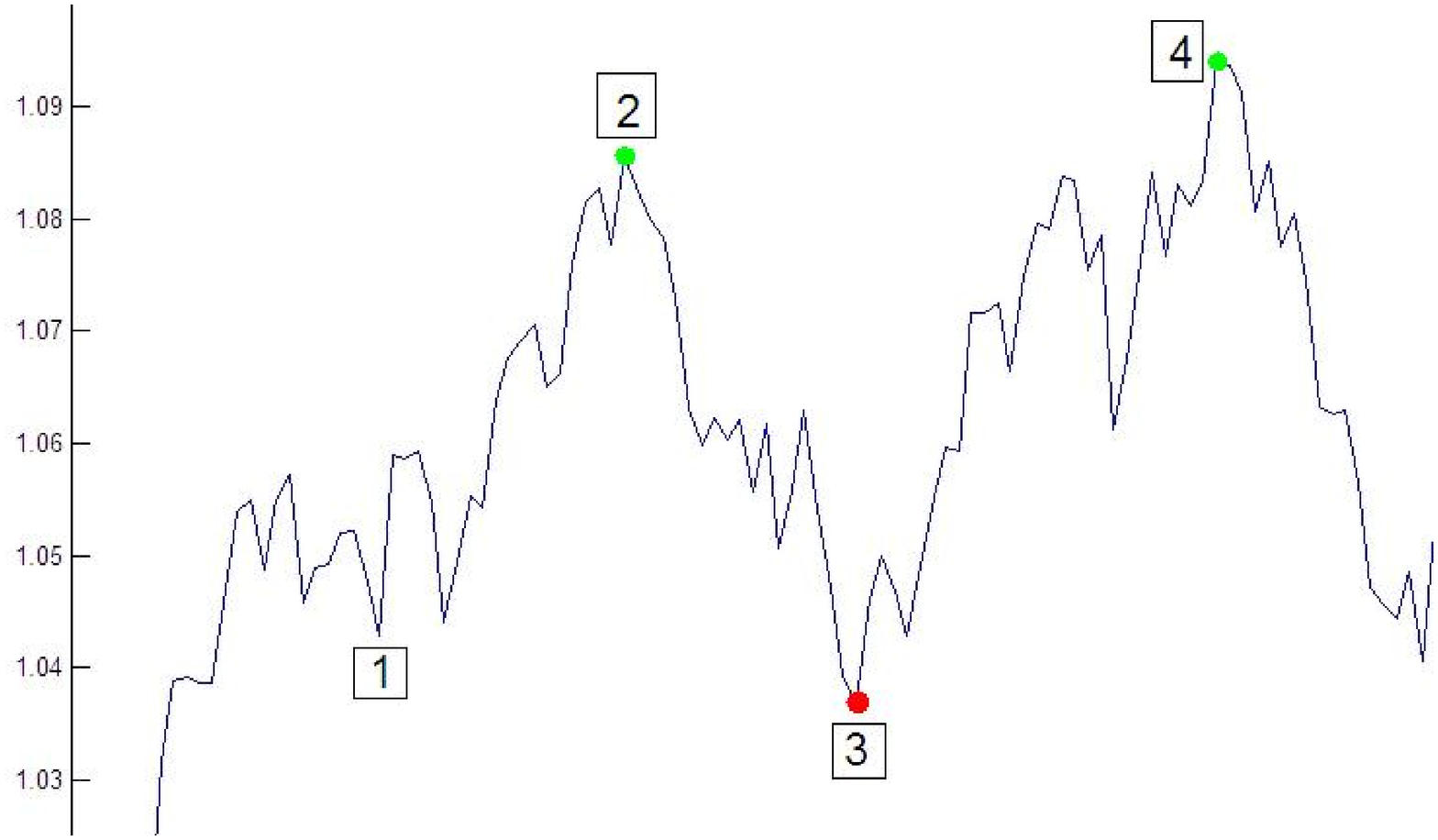}
  }
  
	\subfloat[Pivot turning points of degree 10]
	{
		\label{fig:PivotTurningPoints}
		\includegraphics[width=0.5\textwidth]{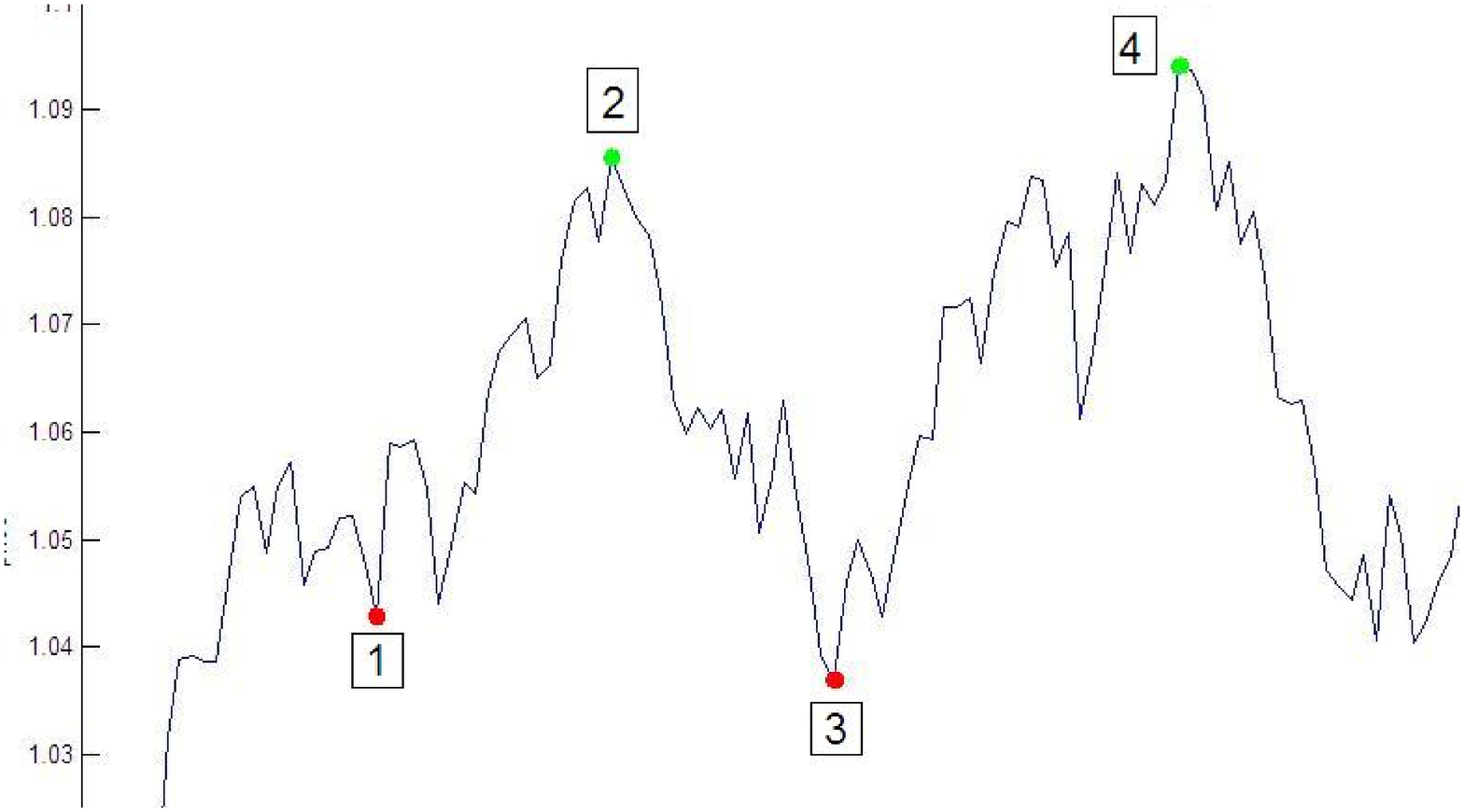}
	}
	\\\subfloat[Turning points of momentum $\gamma=0.03$, lookahead $p=6$]
	{
		\label{fig:VariationTurningPoints}
		\includegraphics[width=0.5\textwidth]{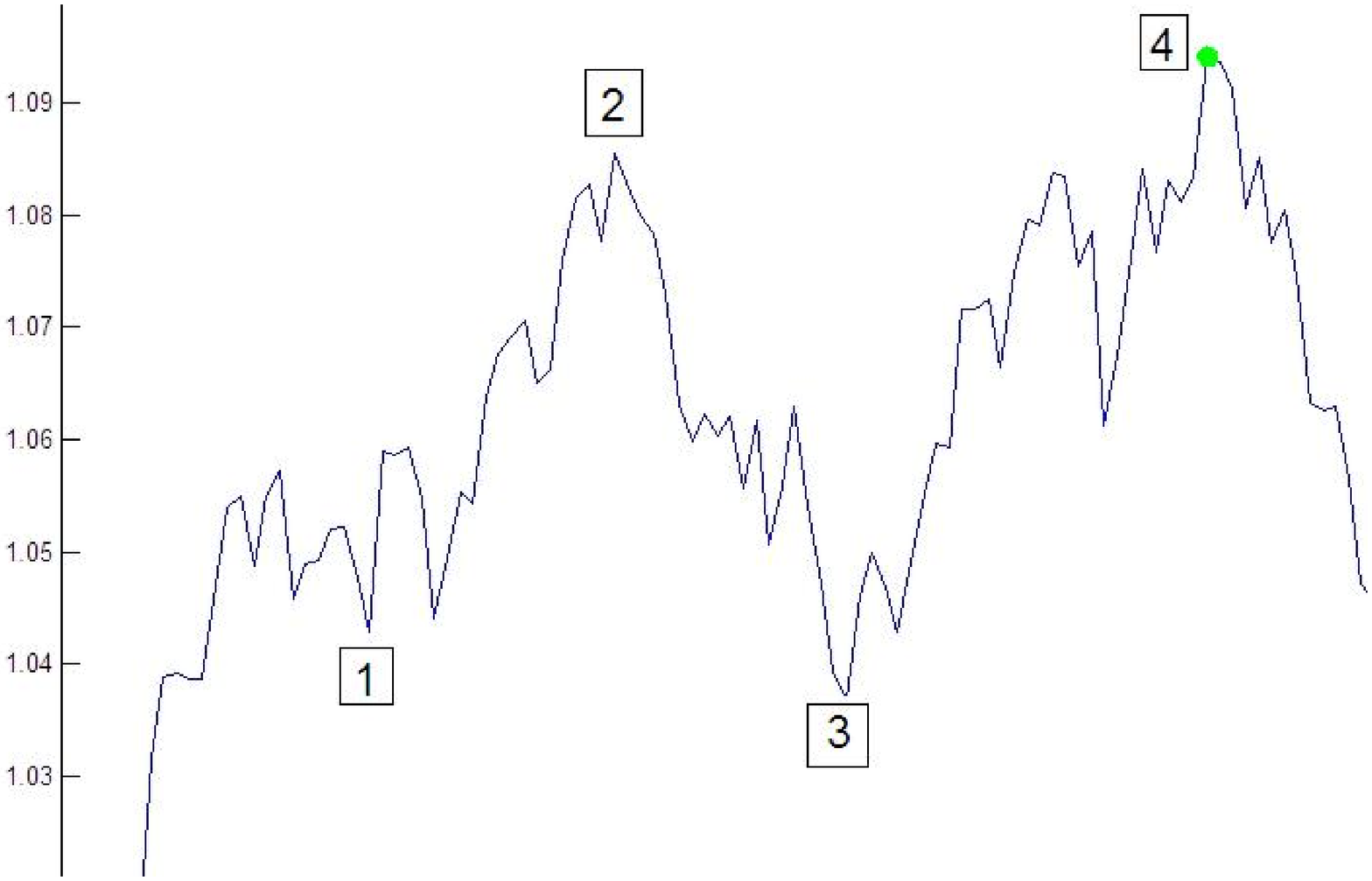}
	}
	\caption{Examples of turning points for different pivot requirements. The turning points that satisfy a requirement are denoted with circles}
	\label{fig:TurningPointsDefinitions}
\end{figure}

When extracting alternating pivot sequences, different requirements will, of course, result in different pivot sequence. 
Figure~\ref{fig:TurningPointsDefinitions} depicts the resulting alternating pivot sequences for three different pivot requirements,
when the input sequence $X$ consists of 100 days the Dow Jones Industrial Average (DJIA) index 
from 05/11/2004 to 1/4/2005.

\subsection{A turning points oscillator}
\label{sec:TurningPointOscillator}

Unlike typical regression problems, where one is interested in predicting prices themselves,
when considering turning points, it is not clear at the outset what should be the
target function. Here we adapt the solution proposed by Li et al. model \cite{LiDL09a}. 
Fixing a class of pivot points (satisfying any desired characteristic), the idea is to construct 
an ``oscillator,'' whose swings correspond to price swings and its extrema points corresponds to
our turning points in focus. 
The oscillator essentially normalizes the prices so as to assign the 
same numerical value (0) to all troughs and the same value (1) to all peaks.
This oscillator provides the target function to be predicted in our regression problem. 
The construction of the oscillator is based on alternating pivot sequences as discussed in Section~\ref{sec:altPivots}.
Throughout the paper, whenever we consider a pivot sequence, the meaning is that we refer to pivots of a certain type
without mentioning this type. In the empirical studies that follow, this type will always be specified.

\begin{definition}[Turning point oscillator (TP Oscillator)]
\label{def:TurningPointOscillator}

Let $X$ be a price sequence and let $A(X)$ be its alternating pivots sequence (that is, $A(x)$ is the list of turning point times in $X$).
The \emph{TP Oscillator} is a mapping $\Gamma: \mathbb{N} \to [0,1]$,
				  		\begin{equation}
				  		\Gamma(t) = \begin{cases}
								 0,  \quad \mbox{if } t  \mbox{ is a trough}, \\
								 1, \quad \mbox{if } t  \mbox{ is a peak}, \\
								 \frac{x_t-P(t)}{P(t)-T(t)}, \quad \mbox{ otherwise}, \\
				  		\end{cases}
				  		\end{equation}
				  		 where P(t) and T(t) are the values of the time series at the nearest peak and trough located in opposite sides of time t.
  \end{definition}

Notice, that for each time index $t$, the TP Oscillator represents the degree of proximity of the price $x_t$ to the price at the nearest peak or trough. 
  Prices that are closer to troughs will have lower values and prices that are closer to peaks will have higher values. 
  The TP Oscillator is clearly bounded in the interval $[0,1]$ and attain the boundary values at troughs and peaks.
 
\begin{figure}[ht]
  \centering
	\subfloat[$\gamma=0.01$]{\includegraphics[width=0.5\textwidth]{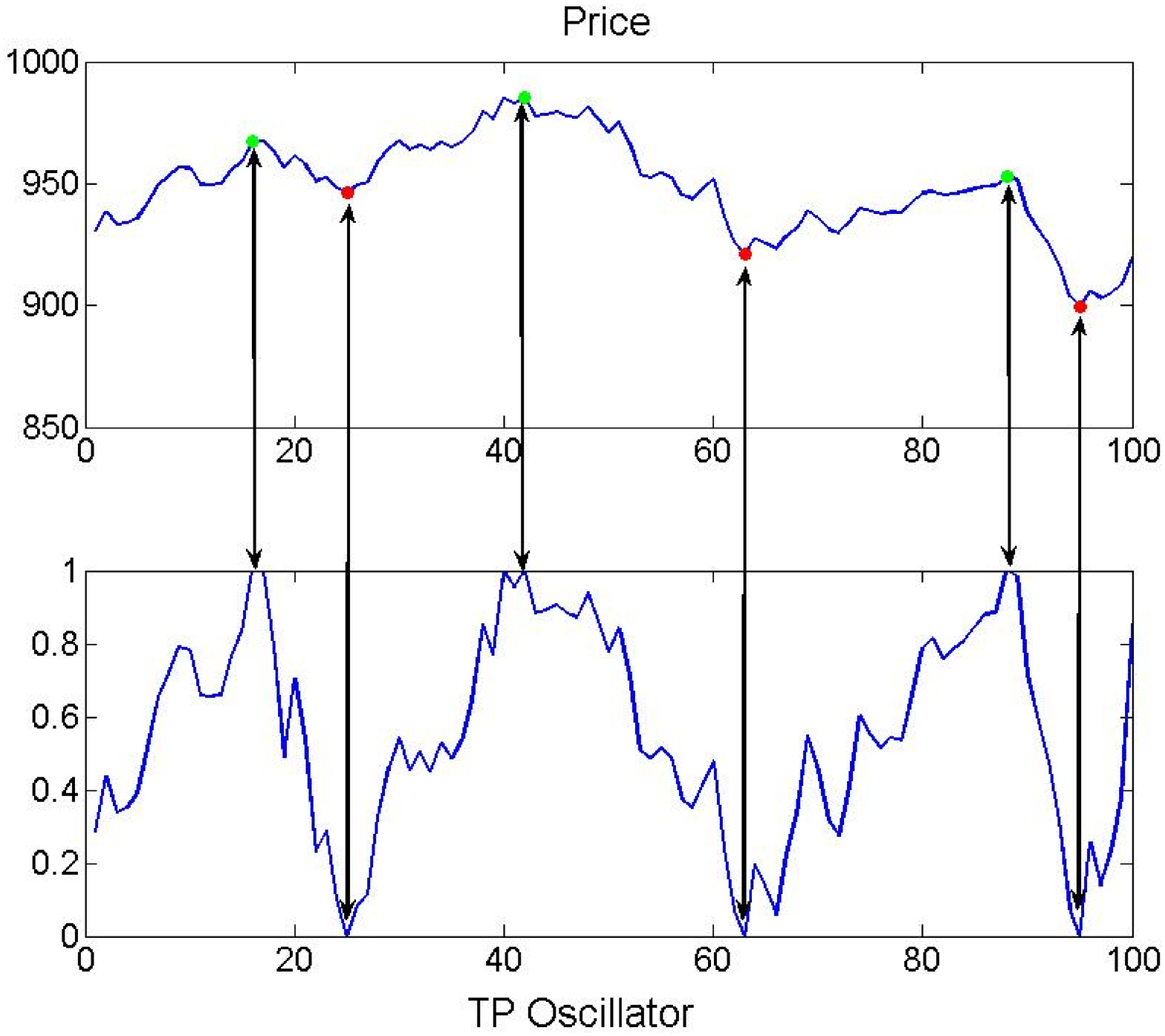}\label{fig:1percent}}
	\subfloat[$\gamma=0.05$]{\includegraphics[width=0.5\textwidth]{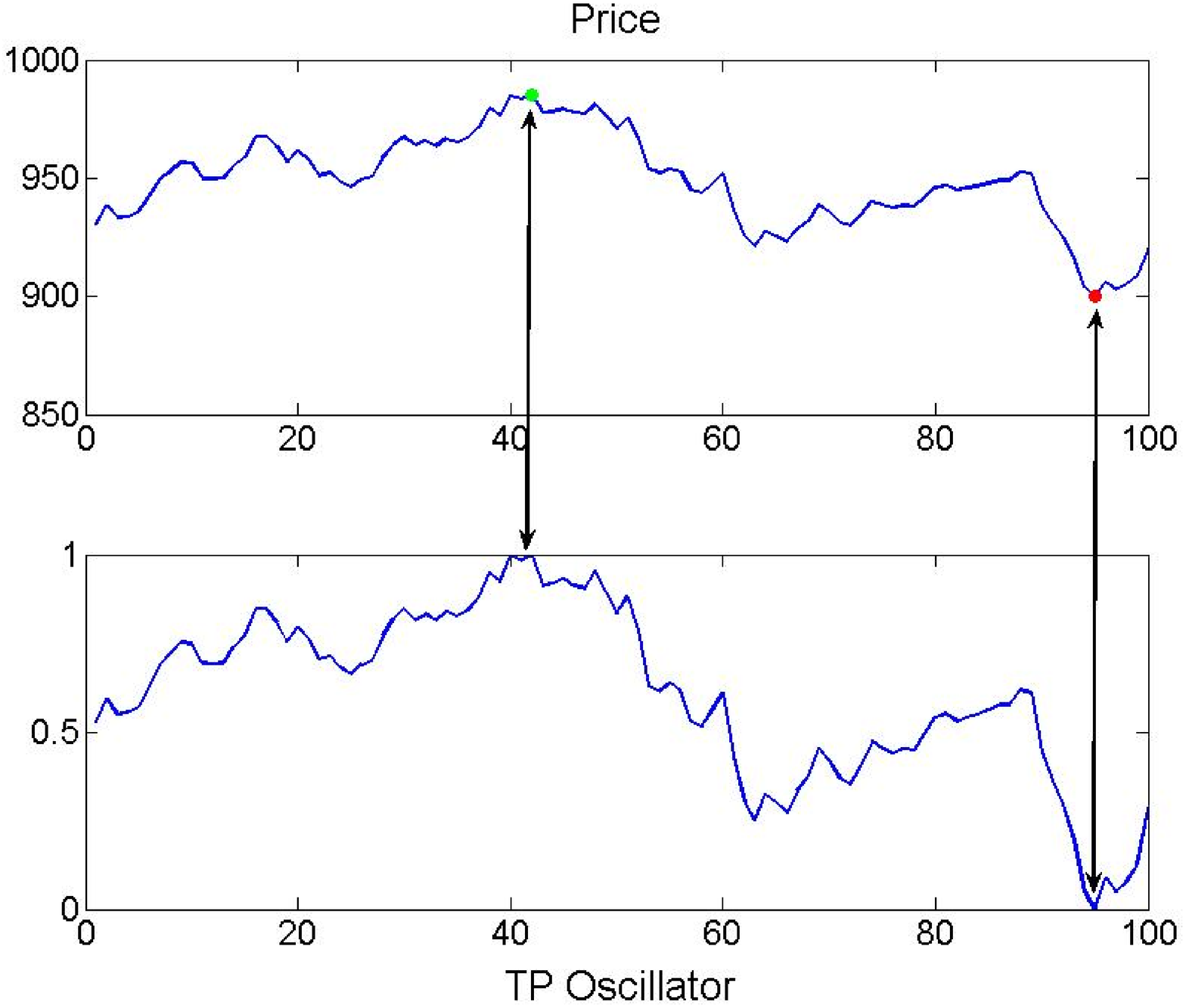}\label{fig:5percent}}
	\caption{TP oscillator for impact turning points}
	\label{fig:TurningPointOscillatorExample}
\end{figure}

 The structure of the TP oscillator strongly depends on the type of turning points used in its construction.
  For instance, in Figure~\ref{fig:TurningPointOscillatorExample} we see examples of the TP Oscillator computed for turning points with impact 
  $\gamma=1\%$ (\ref{fig:1percent}), and impact $\gamma=5\%$ (\ref{fig:5percent}). As should be expected, there are more peaks and troughs in 
 Figure~(\ref{fig:1percent}) than in Figure~(\ref{fig:5percent}) because the number of pivot points of smaller impact is larger,
 so the TP oscillator attains its extreme values more frequently. 

\section {On the choice of features for turning points prediction}
\label{sec:featuresChoice}
In many prediction problems the choice of features is a crucial issue with tremendous impact on performance.
From a learning theoretic perspective this choice should be done in conjunction with the choice of the model class.
The overall representation (features plus model class) determines learnability and predictability. 
Since we focus in this paper on autoregressive models, the features we consider are limited to multivariate functions of past data.

In the Li et al.  paper \citep{LiDL09a}, on which we build, it was advocated that price evolution is the outcome of a nonlinear chaotic dynamics 
and therefore, they used tools from chaos theory to determine the length of the  \emph{backward window} of prices from which to 
generate the features. The features themselves where simply normalized prices.
Specifically,  based on Takens theorem \citep{Takens:10.1007/BFb0091924}, the length of this backward window was derived 
as the minimum embedding dimension of the training price sequence.
The TP Oscillator used by Li et al. was defined in terms of momentum turning point. Recalling Definition~\ref{def:SpeedTurningPoint}, 
momentum pivots are characterized via two parameters: the `importance' parameter $\gamma$, and the lookahead window length. 
Li et al. focused on short term oscillations, and the choice of the corresponding
lookahead window length was based on the \emph{Lyapunov exponent}.\footnote{
In chaotic dynamical systems, the horizon of predictability, which is directly affected sensitivity to initial conditions,
is inversely proportional to the maximal Lyapunov exponent.}
It is perhaps surprising that their chaotic dynamic analysis resulted in a conclusion  
that the past eight prices are sufficient for predicting the turning points.

We decided not to take the Li et al. choice of representation for granted and performed an initial study where we 
considered several types of features and backward window lengths. 
In our study we employed a `wrapper' approach \cite{KohaviJ97} where we
quantified the performance of the features within the entire system (so that trading performance determined the quality of the features).

The conclusion of our study was that the better performing features (among those we considered) are normalized prices within a backward window
as well as the Fourier coefficients of those prices. The Fourier coefficients were simply the phase and amplitude coefficients resulting from
a standard application of the discrete Fourier transform over the backward window.
Our initial study confirmed that the better performing backward window length is eight, exactly as  the Li et al. conclusion.

\section {Predicting turning points with Support Vector Regression}
\label{sec:predictWithSVR}

In this section we describe our application of Support Vector Regression (SVR) to predict the TP oscillator
(presented in Section~\ref{sec:TurningPointOscillator}).
We refer the unfamiliar reader to \cite{Vapnik:1995:NSL:211359} and \cite{Smola:2004:TSV:1011935.1011939} for comprehensive expositions of SVR. 

The SVR application is as follows.
We train an SVR to predict the TP Oscillator $\Gamma(t)$ as a function of the features, which are extracted from price information contained in 
the most recent backward window. Denote by $F: \mathbb{R}^m \to \mathbb{R}^d$ the feature generating transformation from prices in the backward window
$W_t = X_{t-m}^{t-1}$ to the feature space. Thus, $F(W_t)$ is a $d$-dimensional feature \emph{vector}. The specific 
transformation $F$ we used, consisted of normalized prices combined with normalized Fourier coefficients as described in Section~\ref{sec:featuresChoice}.
The label corresponding to the feature vector $F(W_t)$ of the $t^{th}$ backward window, 
is set to $\Gamma(t)$, the TP Oscillator at time $t$.
This way we consider a training set $S_n = (F(W_t),\Gamma(t))$, $t_1 \leq t \leq t_2$, of $n = t_2 - t_1 +1$ consecutive pairs of features with their labels.

Using $S_n$ we train a support vector regression model. 
To this end we use an $\epsilon$-SVR model (and a training algorithm) \citep{CC01a} using a radial basis function (RBF) kernel,
$K(x_i,x_j)=\exp ({-\frac{1}{\sigma^2}\|x_i-x_j\|^2})$. 
The model is controlled by three hyper-parameters $C$, the error cost, $\epsilon$ the tube width, and $\sigma$, the kernel resolution. 
The reader is referred to \cite{Smola:2004:TSV:1011935.1011939} for a discussion on the role of these parameters.
The output of the SVR training process is a 
function $\hat{\Gamma}(t): \mathbb{R}^d \to \mathbb{R}$, from feature vectors to the reals.
In our case, the function $\hat{\Gamma}$ is the SVR functional approximation to the TP oscillator, induced from the training set.
Thus, larger values of $\hat{\Gamma}$ reflect closer proximity to peaks, 
and smaller values, closer proximity to  troughs.

Since $\hat{\Gamma}(t)$ is only an approximation of $\Gamma(t)$ reflecting relative proximity to extrema points,
we cannot expect that its own extrema points will explicitly identify the peaks and troughs themselves.
Therefore, in order to decide what are the magnitudes of the predicted values that should be considered as peaks or troughs, 
we introduce thresholds $T_{low}, T_{high}$ so that indices $t$ such that $\hat{\Gamma}(t) < T_{low}$, are all treated as troughs, 
and conversely, indices $t$ for which  $\hat{\Gamma}(t) > T_{high}$, are all considered as peaks. The thresholds $T_{low}$ and $T_{high}$ 
are hyper-parameters that are part of our model and should be fitted using labeled training data.

\subsection {Problem specific error function and optimization}
\label{sec:CustomFunction}

Various error functions are used in regression analysis. The most common ones are
the root mean square error (RMSE) and the mean absolute error (MAE).
For example, the RMSE of the prediction $\hat\Gamma(t)$ over the subsequence $X_{t_1}^{t_2} = x_{t_1},\ldots,x_{t_2}$ of  $n=t_2-t_1 + 1$ inputs, is given by,
$$
RMSE = RMSE(\hat\Gamma(t),\Gamma(t),X_{t_1}^{t_2}) = \big(\frac{1}{n}\sum_{t=t_1}^{t_2} (\hat\Gamma(t)-\Gamma(t))^2\big)^{\frac{1}{2}}.
$$
Instead of directly using the RMSE, Li et al. \citep{LiDL09a} suggested to use a problem specific variant of the RMSE.
This specialized error function is defined in terms of the following trimmed reference function,
$$\Gamma^{'}(t) \eqdef
		\begin{cases}
			T_{high}, \ \ \ \mbox{if } \Gamma(t)=1 \mbox{ and } \hat{\Gamma}(t)<T_{high} & \mbox{(high false negative)}\\
			T_{high}, \ \ \ \mbox{if } \Gamma(t) \ne 1 \mbox{ and } \hat{\Gamma}(t)>T_{high} & \mbox{(high false positive)} \\
			T_{low}, \ \ \ \mbox{if } \Gamma(t)=0 \mbox{ and } \hat{\Gamma}(t)>T_{low} & \mbox{(low false negative)}\\
			T_{low}, \ \ \ \mbox{if } \Gamma(t) \ne 0 \mbox{ and } \hat{\Gamma}(t)<T_{low} & \mbox{(low false positive)} \\
			\hat\Gamma(t), \ \ \ \mbox{otherwise.}
	  \end{cases}
$$				

\begin{figure}[htbp]
	\includegraphics[width=0.85\textwidth]{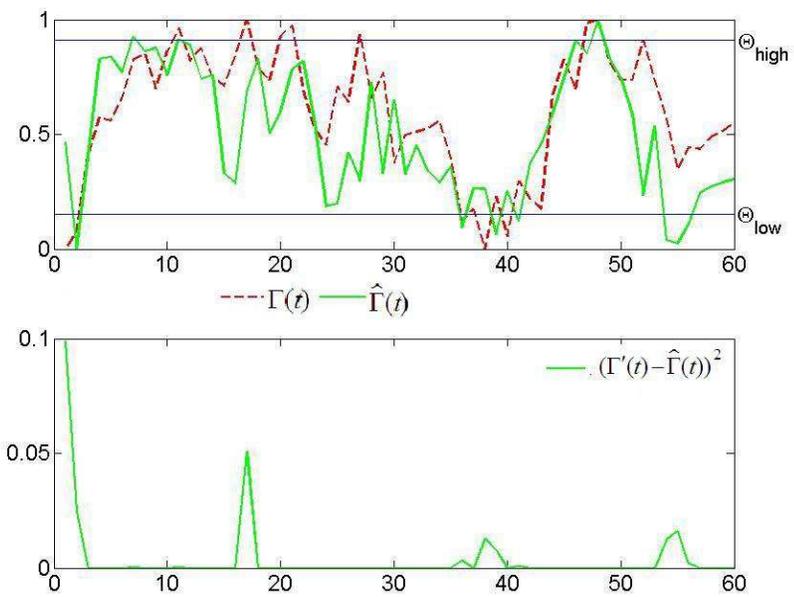}
	\caption{Calculation of TpRMSE. The upper graph shows the values of the actual TP Oscillator $\Gamma(t)$ and the predicted TP Oscillator $\hat\Gamma(t)$. The lower graph identifies regions that actually influence the error magnitude, expressed at each point as $(\Gamma'(t)-\hat\Gamma(t))^2$.}
	\label{fig:TpRMSEExample}
\end{figure}

The final specialized error function, denoted $TpRMSE $, is then defined in terms of the reference function as,
$$
TpRMSE \eqdef TpRMSE(\Gamma(t), \hat{\Gamma}(t), T_{low}, T_{high}) \eqdef \big(\frac{1}{n}\sum_{t=t_1}^{t_2} (\hat{\Gamma}^{'}(t)-\hat\Gamma(t))^2\big)^{\frac{1}{2}}.
$$ 
We observe that $\Gamma^{'}(t) - \hat\Gamma(t)$ has non-zero values at values $t$ corresponding to
wrong predictions, thus allowing to form an error function that penalizes such deviations according to their magnitude. 
An example shown the TpRMSE error function is given in Figure \ref{fig:TpRMSEExample}.

\begin{remark}
Li et al. \cite{LiDL09a} proposed to use the following refinement of the $TpRMSE$ error function, which assigns larger weights to errors occurring precisely at true turning points, and smaller weights for other errors,
$$
\TPRMSEBETA \eqdef \big(\frac{1}{n}\sum_{t=t_1}^{t_2} \beta_{t}(\hat{\Gamma}^{'}(t)-\hat\Gamma(t))^2\big)^{\frac{1}{2}},
$$
The exact choice of the $\beta$ coefficients can be found in \cite{LiDL09a}. 
In our experiments, whenever we applied the Li et al. model we used this refined $\TPRMSEBETA$ error function.
However, we did not find any significant advantage to this refined error function and
our SVR model was applied with the simpler $TpRMSE$ error function.
\end{remark}

\noindent
Given a fixed SVR regression estimate $\hat{\Gamma}(t)$, we need to optimize the thresholds $T_{high}$ and $T_{low}$, so as to minimize 
the $TpRMSE$ error. 
The optimal thresholds are thus given by, 
\begin{equation}\label{eq:thresholdOptimize}
[T_{low}, T_{high}] = \argmin_{T^{'}_{low}, T^{'} _{high}} TpRMSE(\Gamma(t), \hat{\Gamma}(t), T^{'}_{low}, T^{'}_{high}).
\end{equation}

Our overall turning point SVR model is specified by the following hyper-parameters. The SVR hyper-parameters are $C$, 
$\epsilon$ and $\sigma$,  and the turning point identification hyper-parameters are $T_{low}$ and $T_{high}$. 
These hyper-parameters need to be optimized over the training sequence.
To this end, we split the training sequence $X$ into two segments, one for optimizing the SVR parameters and the other, for optimizing the thresholds (in the sequel we call this second segment of the training sequence the \emph{validation segment}).
Due to the complexity of the error functional we resorted to the exhaustive grid search.
Specifically, both sets of hyper-parameters were selected exhaustively over appropriate set of values (a grid),
so as to minimize the error functional. 
The SVR grid is denoted
$\Theta(C,\sigma,\epsilon)$ and the thresholds grid is denoted 
$\Theta(T_{low}, T_{high}$). In Section \ref{sec:SVRDetails} we discuss the particular choices of these grids.
The overall model selection strategy is summarized in Algorithm \ref{alg:ModelProducing}.

\begin{algorithm}
	\caption{SVR TPP model}
	\label{alg:ModelProducing}
	\begin{algorithmic}
	\FORALL{$(C,\sigma,\epsilon) \in \Theta(C,\sigma,\epsilon)$} 
		\STATE Train an SVR model on the training data segment 
		\STATE Optimize $(T_{low}, T_{high}) \in \Theta(T_{low}, T_{high})$ on the validation segment using \eqref{eq:thresholdOptimize}
		\STATE Store $TpRMSE$ and $(T_{low}, T_{high})$ for the given triple $(C,\sigma,\epsilon)$
	\ENDFOR
	\STATE Select the $(C,\sigma,\epsilon)$ obtaining the minimal TpRMSE score on the validation sequence with their corresponding
	optimal thresholds $(T_{low}, T_{high})$.
	\end {algorithmic}
\end{algorithm}

We note that Li et al. \cite{LiDL09a}  applied a genetic algorithm to optimize the corresponding thresholds 
($T_{low}$, $T_{high}$) in their model. 
However, we found that the exhaustive search over a grid performs the same or better than a genetic algorithm,
with tolerable  computational penalty. The main advantage in using our deterministic approach, rather than
the randomized genetic algorithm, is the increase of reproducibility (our optimization is deterministic 
and always has the same outcome).

\section{A trading application}
\label{sec:TradingStrategy}
Prediction algorithms can be evaluated through any meaningful error function. For example, we could evaluate 
our methods using the TpRMSE function defined above. However, the degree of meaningfulness of an error function depends on an application.
Perhaps the most obvious application of turning points predictions is trading. 
We now describe a very simple trading strategy implementing the infamous \emph{buy low sell high} adage. 
This strategy will be used to assess performance of our method, including comparisons to 
a natural benchmark and to the state-of-the-art model of \cite{LiDL09a}.

The idea is to use predictions from our already constructed regressor, $\hat{\Gamma}$, of the TP Oscillator, $\Gamma$, 
to issue buy and sell signals. 
  Given the prediction $\hat{\Gamma}(t)$ and the thresholds $T_{low}$ and $T_{high}$, trading 
  operations are triggered according to the following rule.
  $$
  Trigger(t) = \begin{cases}
								 Buy,  \ \ \ \mbox{if }\hat{\Gamma}(t) < T_{low}  \mbox{ and not in position} \\
								 Sell, \ \ \ \mbox{if }\hat{\Gamma}(t) > T_{high}  \mbox{ and in position}
				  		\end{cases}
  $$
The strategy thus works as follows. If today we are not in position (i.e., we are out of the market) and a buy signal is
 triggered at the market close, we enter a long position first thing tomorrow, on the opening price. 
 We start this trade by buying the stock using our entire wealth.
As long as we are in the trade, we wait for a sell signal, and as soon as it is triggered, we clear our position on the opening of the following day.
(i.e., clearing the position means that we sell our entire holding of the stock).

To evaluate trading performance we will use \emph{cumulative return}, \emph{maximum drawdown}, \emph{success rate} and the 
\emph{Sharpe ratio} measures. These are standard quantities that are often used to evaluate trading performance.
To define these measures formally, let  $x_t^{t+n}$ be a price sequence of length $n$, and let 
$\{(b_i, s_i), b_i, s_i \in [t, t+n], i=1,\ldots,L\}$, be $L$ pairs of times corresponding to matching buy and sell ``triggers'' generated by a trading strategy $S$. This pairs correspond to $L$ buy/sell trades. 
Thus, for any $i$, we bought the stock at price $x_{b_i}$ and sold it at price $x_{s_i}$.

The \emph{Cumulative return}, $\CRS(x_t^{t+n})$, of a trading strategy $S$ with respect to the price sequence $x_t^{t+n}$, 
is  the total wealth accumulated at the end of a test period,
 assuming full reinvestment of our current wealth in each trade, and starting the test with one dollar,
\begin{equation}
\label{eq:cumulativeReturn}
	\CRS(x_t^{t+n}) = \prod {\frac{x_{s_i}}{x_{b_i}}}.
\end{equation}	
	The corresponding \emph{annualized cumulative return}, $\ACRS(x_t^{t+n})$ is,\footnote {There 252 business days in a year.}
\begin{equation}	
	\ACRS(x_t^{t+n}) = [\CRS(x_t^{t+n})]^{252/n}.
\end{equation}
	
It is often informative to consider also the \emph{rolling} cumulative return, $\ROCS(t)$, which is the curve of cumulative return through time,
\begin{equation}
	\ROCS(t) = \begin{cases} 
							  					1 \mbox{, if } t \notin \bigcup([b_i,s_i]) \\
												  x_t/x_{t-1} \mbox{, if } t \in \bigcup([b_i,s_i]) \\
												 \end{cases}, t=t+1, \ldots, t+n	
\end{equation}

\begin{figure}[htbp]
	\centering
		\includegraphics[width=0.50\textwidth]{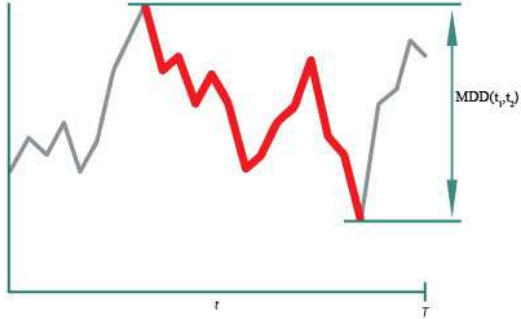}
	\caption{Maximum drawdown (MDD) example}
	\label{fig:MDD}
\end{figure}

The \emph{maximum drawdown (MDD)}, $\MDDS(x_t^{t+n})$, is a measure of risk, defined with respect to the rolling cumulative return curve.
Consider Figure \ref{fig:MDD}, depicting a hypothetical (rolling) cumulative return curve.
The MDD of this curve is defined to be the maximum cumulative loss from a peak to the following minimal trough during a trading period. 
The MDD is emphasized in the Figure~\ref{fig:MDD} by the measured height. 
Formally, if $\ROCS(k)$ is a the cumulative return sequence, its MDD in the time interval $[t, t+n]$ is,
\begin{equation}
\MDDS(x_t^{t+n}) = \max_{\tau \in [t,t+n]}\{\max_{k \in (t, \tau)} [\ROCS(k) - \ROCS(\tau)]\}.
\end{equation}

The \emph{Sharpe ratio}(SR), $\SR(x_t^{t+n})$, is a risk-adjusted return measure \citep{Sharpe1994}. Intuitively it characterizes how smooth and steep
the rolling cumulative return curve is,
\begin{equation}
\SR(x_t^{t+n}) = \frac{\CRS(x_t^{t+n})-1}{std(\ROCS(x_t^{t+n}))}.
\end{equation}	

It is common to annualize SR in order to be able to compare SR for different time periods. Thus, for daily data, the
 \emph{annualized Sharpe Ratio (ASR)} is,
\begin{equation}
\ASR(x_t^{t+n}) = \frac{\ACRS(x_t^{t+n})-1}{std(\ROCS(x_t^{t+n})) \cdot \sqrt{252} }.
\end{equation}	

The last metric we introduce is the \emph{rate of success}, $\RATES_S(x_t^{t+n})$, which is simply 
proportion of successful trades,
\begin{equation}
\label{eq:successRate}
\RATES_S(x_t^{t+n}) = \frac{\sum{\I(x_{s_i} > x_{b_i})}}{L}.
\end{equation}

As a natural benchmark for trading performance we take the \emph{buy-and-hold (BAH)} strategy, which simply buys the stock 
at the beginning of the period and then sells it in the end of the period.
While this strategy is very simple (and perhaps naive), it makes much sense 
in the stock market.\footnote{In fact, many economists believe that BAH is the only strategy that makes sense \cite{ Malkiel1973}.}
Calculations of the above performance metrics for the BAH strategy are straightforward, according
to the above formulas (\eqref{eq:cumulativeReturn}) - (\eqref{eq:successRate}) by assigning $b_1 = t, s_1=t+n$, and $L=1$.

\subsection {Example: predicting turning points with SVR}
\label{sec:exampleSVR}
\begin{figure}[htbp]
	\centering
	\subfloat[TP Oscillator and its prediction]
	{
		\includegraphics[width=0.80\textwidth]{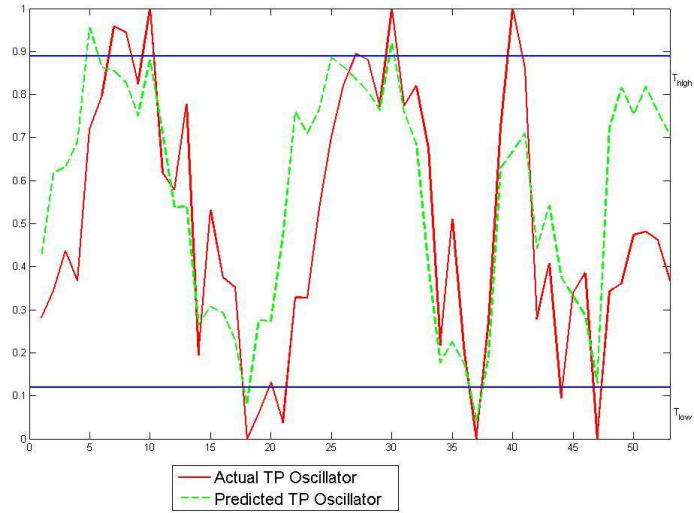}
		\label{fig:MethodExample}
	}
	
	\qquad
		
	\subfloat[Applying TP Oscillator and thresholds for TP prediction. The upper chart depicts prices, circles mark the actual turning points and squares mark predicted turning points. The lower chart depicts the TP Oscillator prediction, the horizontal lines are the thresholds.]
	{
	\includegraphics[width=0.80\textwidth]{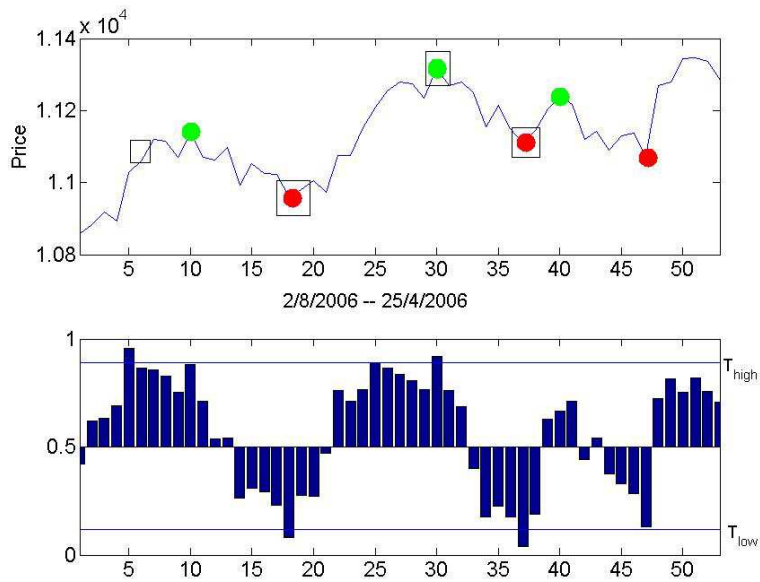}
	\label{fig:StrategyExample}
	}
	\caption{Demonstration of SVR model predictions over an out-of-sample DJIA data}
\end{figure}

In Figure \ref{fig:MethodExample} we demonstrate predictions obtained by our SVR model over an out-of-sample (i.e., test) data. 
In this example we restricted attention to turning points with impact $\gamma=0.01$, and trained the model
over one year of the DJIA index from 07-Sep-2004 to 09-Nov-2005. The SVR hyper-parameters and the thresholds were selected over 
a validation segment from 11-Nov-2005 to 30-Jan-2006, and predictions were performed for an out-of-sample test period 
from 08-Feb-2006 to 25-Apr-2006 (this period is shown in the figure).
 	In this example feature vectors consisted of the past eight prices normalized to reside in $[0,1]$. 
The SVR hyper-parameters were optimized over the grid 
$\Theta(C,\sigma,\epsilon) = 
[2^{-10}, 2^{-9}... 2^{10}]\times[2^{-10}, 2^{-9}... 2^{10}]\times[0.01, 0.05, 0.1]$. 
The best model parameters that were found in this grid were $C = 32, \gamma=0.0625, \epsilon = 0.05$.
The turning points identified by this model are marked in the figure.
We note that the prediction problem (data and train/test periods) in this demonstration is precisely the same as
the one used in \cite{LiDL09a} for evaluation.

\section{Predicting turning points with neural networks}
\label{sec:AnnTPPDescription}

In this section we briefly describe the original turning points prediction model of Li et al.  \cite{LiDL09a}, which relies on neural networks.
In their methods, the feature vector consists of normalized prices within a recent backward window. The size of this window is
determined by calculating the embedding dimension over the training sequence.
The target function to be regressed and predicted is the TP Oscillator restricted to turning points with momentum
$\gamma$ and lookeahead window length which is calculated using the Largest Lyapunov exponent of the training sequence
(see details in \cite{LiDL09a}).
The training set is constructed as described in Section~\ref{sec:randomSegments}.
The model trains a number of neural networks, which differ in the number of neurons in their single hidden layer.
These networks form an ensemble whose prediction is generated by aggregating individual networks outputs using
their weighted sum.
These ensemble weights, as well as the prediction thresholds, are optimized to minimize the 
$\TPRMSEBETA$ function (see Section~\ref{sec:CustomFunction}) over a validation sequence.
For further details on how the backward window size is determined, the reader is referred to \cite{LiDL09a}. 
We only note that the empirical studies conducted in \cite{LiDL09a} considered two price sequences, DJIA including 411 points, and TESCO consisting of 492 points.
In both cases the backward window length was taken to be eight days.

\section {Comparison between the SVR and the ANN methods}

Having described the SVR and the ANN models in Sections~\ref{sec:predictWithSVR} and~\ref{sec:AnnTPPDescription}, respectively, we would like to emphasize 
the resemblance and differences between these two methods.
Both methods are focused on predicting turning points and in general, we were motivated by the Li et all. paper \citep{LiDL09a},
followed their general outline,
and drew on a number of their ideas, including the utilization of their TP oscillator and 
their loss function.
Our variant differs in several aspects.
Aside from the different regression algorithm (their ensemble of neural networks vs. our support vector regression), 
our model relies on a different feature set and utilizes a different target function.
Specifically, the feature set in \citep{LiDL09a} is a window of the past prices. In our case, the features are both prices and the
Fourier coefficients of the prices. The target function used in \citep{LiDL09a} is the TP Oscillator restricted to
 momentum turning points of particular magnitude (and lookahead window length).
In our case it is the TP Oscillator of defined with impact turning points of particular magnitude. 
The differences between our approach and theirs are summarized in Table \ref{tab:ANNSVRDiff}.

\begin{table}[htb]
\scriptsize	
	\centering
		\begin{tabular} {lll}
		\toprule
		& \cite{LiDL09a} & Our work\\
		\midrule
		Feature vector length & Chaotic embedding dimension & Optimized length (hyper-parameter)\\
		Turning point property & Momentum & Impact \\
		Regression algorithm & Ensemble of neural networks & Support Vector Regression (SVR)\\
		Input representation & Raw prices & Raw prices and Fourier coefficients\\
		\bottomrule
		\end{tabular}
	\caption{Summary of differences between ANN and SVR models}
	\label{tab:ANNSVRDiff}
\end{table}

\section{Experimental design}

We conducted an extensive set of experiments to evaluate the SVR and ANN turning points prediction models
discussed in Sections~\ref{sec:predictWithSVR} and \ref{sec:AnnTPPDescription}, respectively.
Our evaluation is performed in terms of the simple trading application introduced in Section~\ref{sec:AnnTPPDescription}.
We evaluated the models over historical price segments of the Dow Jones Industrial Average (DJIA).
Since financial sequences such as the DJIA exhibit great many behaviors and are extremely noisy, we considered in our experiments many 
predicting tasks, corresponding to many sequence segments along the DJIA history.
To facilitate the discussion we first define and discuss in the following subsections the essential technical aspects in our experimental design.

\subsection {Train, validation and test splits}
\label{sec:randomSegments}
\begin{figure}[htbp]
  \centering
	\includegraphics[scale=0.7]{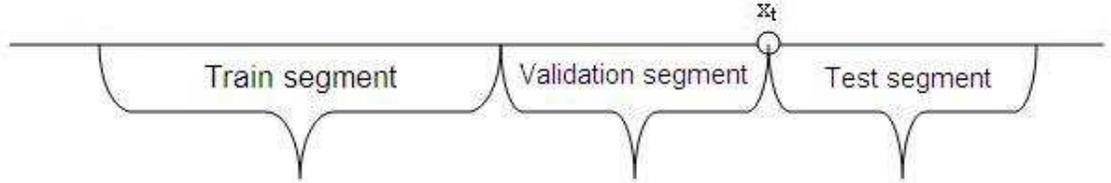}
	\caption{Splitting data into train, validation and test segments}
	\label{fig:datasetSplit}
\end{figure}

Each instance of a prediction task is a triplet of price subsequences of the financial sequence in question (DJIA) 
consisting of \emph{training}, \emph{validation} and \emph{test} segments,
as depicted in Figure~\ref{fig:datasetSplit}. Thus, the prices in each of the segments are contiguous and the segments follow each other chronologically. 
The training segment is used to fit model parameters (for the SVR or ANN models). 
The validation segment is used for model selection via hyper-parameter tuning, and the test segment is utilized to evaluate performance. 
Each such triplet is called a \emph{prediction task}.
We denote the lengths of the three segments by $N_{train}$, $N_{valid}$ and $N_{test}$, respectively, and the choice of these parameters in our experiments will be discussed later. 

\subsection {Statistical validity}

Our tests consider multiple methods  and multiple prediction tasks.  
To ensure statistical validity of our conclusions, we utilize the following statistical tests.

\subsubsection{Tests for pairwise comparisons}
When comparing the performance of two algorithms, or two instances of an algorithm
(i.e.,  the same algorithm applied with different parameters) over multiple tasks, 
we use the Wilcoxon \emph{Signed-Rank test}, as recommended and described in \cite{Demsar06} Section 3.1.3. For a pair of algorithms of interest, we conduct the test and calculate the $p$-value. If the the performance difference can be accepted at 5\% level, we conclude that one of the algorithms is significantly better than the other (at 5\% significance level); otherwise, we conclude that the algorithms are statistically indistinguishable. 

\subsubsection{Tests for multiple comparisons}
When comparing a group of algorithms (more than two) over multiple tasks
we use the Friedman test to determine the rank of each of algorithm. In this case the
$p$-value indicates whether differences between the algorithms are significant. 
If significant differences are observed, we proceed with a 
post-hoc test using the Bonferonni-Dunn test for pairwise rank comparisons in order to find out the best performing algorithm 
in the group. The precise procedure we follow is described in \cite{Demsar06} Section 3.2.2.

\subsection {Software}
All experiments were conducted under Matlab. For training the ANN models we used the Matlab Neural Network toolbox.
The SVR models were computed using the SVMLib toolbox \cite{CC01a}. 

\subsection {Dataset} 
Following \citep{LiDL09a} our data set consisted of close prices of the DJIA index.\footnote{We downloaded the prices from
Yahoo! Finance, \text{http://finance.yahoo.com}.}
We analyzed a large DJIA segment from
1/1/1960 to 1/1/2010. Spanning 50 years, this sequence contains 12585 prices. In this time range we selected 300 prediction (and trading) problem 
tasks uniformly at random. This quite large number of problem instances was chosen to ensure statistical significance.
In particular, the statistics reported are typically averaged over these 300 instances. 
As discussed in Section~\ref{sec:randomSegments}, each of the 300 tasks is a triplet consisting of consecutive train, validation and test segments.
The lengths of these segments are $N_{train} = 504$ days (two years), $N_{validate} = 60$ days, 
and $N_{test} = 60$ days. 
Finally, we note that the 300 test periods in our prediction tasks sometimes overlap, but in general they are uniformly 
spread along the 50 year period.\footnote{The precise periods selected for these 300 tasks were recorded and can be obtained
from the authors.}

\subsection {Details of the SVR model and its hyper-parameters}
\label{sec:SVRDetails}

Our SVR model is described in Section~\ref{sec:predictWithSVR}.
To actually apply this model, we need to make some choices 
regarding representation. Specifically, we need to choose 
the type of turning points to focus on, either pivot (of a certain degree), impact or momentum, and in each case select desired resolution, controlled by the pivot degree or the parameter  $\gamma$ (see Section~\ref{sec:ProblemSetting}).
The SVR model itself is controlled by three hyper-parameters: $C$, $\sigma$ and $\epsilon$.
The role of these parameters is described in \cite{Smola:2004:TSV:1011935.1011939}.
Turning point identification is achieved using two additional hyper-parameters, $T_{low}$ and $T_{high}$, as described in Section~\ref{sec:TradingStrategy}.
The SVR model selection is performed 
using a straightforward grid search for the best  triplet of SVR hyper-parameters. This SVR parameter space
(grid), denoted $\Theta(C,\sigma,\epsilon)$, was chosen to be
$$
\Theta(C,\sigma,\epsilon) = \{0.1, 1, 100\} \times \{0.1, 1, 100\} \times \{0.01, 0.05, 0.1\}.
$$ 
Similarly, the parameter space (grid) for the thresholds was taken to be 
$$
\Theta(T_{low}, T_{high}) = [0 \mbox{ } 1] \times [0 \mbox{ } 1],\ \ \  \mbox{with step $0.01$}.
$$
These choices were made based on a preliminary rough study on other price sequences before conducting the 
experiments and were not optimized thereafter.


\subsection {Details of the ANN model and its hyper-parameters}
\label{sec:ANNDetails}

The ANN model of Li et al. \cite{LiDL09a} is briefly introduced in Section~\ref{sec:AnnTPPDescription}.  
In order to replicate their model as accurately as possible
we followed all their choices. The parameters for the backpropagation learning algorithm were:
\begin{itemize}
	\item Transfer function: hyperbolic tangent;
	\item Output function: linear function.
	\item Backpropagation learning function: gradient descent with learning rate 0.01.
\end{itemize}
The chaotic characteristics of the time series were used for selecting some of the parameters as follows:
\begin{itemize}
\item The \emph{embedding dimension} $m$ of the training segment was used to determine the backward window length. 
Embedding dimension calculations were performed using Cao's method \cite{Cao:1997:PMD:272840.272844} 
as implemented by Merkwirth et al. in their Matlab toolbox 
\cite{OpenTSTool}.
\item The \emph{time delay} $\tau$ was used to determine the sampling rate of the input data. 
The time delay was calculated in accordance with the mutual information method described in \cite{PhysRevA.33.1134},
and implemented in the Merkwirth et al. Matlab toolbox as well \cite{OpenTSTool}.
\item The largest \emph{Lyapunov exponent} $\lambda$ was used to determine the lookahead window length (required to define
the impact turning points), and was set to $1/\lambda$, as was done in \cite{LiDL09a}. 
This Lyapunov exponent was calculated using Rosenstein et al's method \cite{Rosenstein:1993:PMC:153895.153901} 
as implemented by Hegger et al. in their  TISEAN tool \cite{HEG99}.
All these chaotic parameters were calculated over the training data segment. 
\end{itemize}

Since the neural network training algorithm starts with random initial network weights, we performed standard `random restarts' to initialize these weights.
We used 10 random restarts for the training of each prediction task as is common in practice. 
Thus, we obtained 10 sets of results for each instance and selected the best performing one over the validation segment.

With these methods for parameter selection we were able to reproduce the results of \cite{LiDL09a} 
for the particular prediction task (i.e., a particular training/test segment) used in their paper for evaluation.
(this prediction task is used in our example of Section~\ref{sec:exampleSVR}).
Since Li et al. used only one DJIA task for evaluation,
one of our contributions is a more through analysis of their method using multiple tasks.

\subsection {Results}
In this section we present the results of our experiments. 
Throughout the presentation, SVR refers to the proposed model and ANN refers to the original method of \cite{LiDL09a}.

\subsubsection {Experiment 1: SVR vs. ANN vs. BAH}
\label {sec:SVRvsANN}
	
The first set of results is a comparative study of SVR and ANN. For both methods, we experimented with 
two types of target functions. 
The functions we considered in this study were
generated using either \emph{momentum pivots} (as originally proposed in \cite{LiDL09a})
of varying degrees, or \emph{impact pivots} of varying degrees. 
The other parameters of these experiments are summarized in Table \ref{tab:SVRvsANN}.
	\begin{table}[htbp]
	\footnotesize
	\centering
		\begin{tabular}{rcc}
			\toprule
			& ANN & SVR \\
			
			\midrule
			Features & Raw prices & Raw prices and Fourier coefficients \\
			Input length & Chaotic &  8 \\
			Time delay   & Chaotic &  1 \\
			Look-ahead   & Chaotic &  - \\\
			$N_{train}$ (training segment length) & 2 years & 2 years \\
			$N_{validate}$ (validation segment length) & 60 days & 60 days \\
			$N_{test}$ test segment length & 60 days & 60 days \\
			\bottomrule
		\end {tabular}
		
	\caption {Experiment 1: Parameters of the SVR and ANN models. The specification `Chaotic' in the ANN
	column means that the parameters
	were selected based on chaotic dynamics analysis as described in Section~\ref{sec:ANNDetails}.}
	\label{tab:SVRvsANN}
	\end{table} 
	
The results of the experiments are presented in Table \ref{tab:TPPvsSVRresults}.
The table summarizes the performance obtained with the two types of pivot points 
(momentum and impact).
For each pivot type we show the performance for a number of $\gamma$ values (recall that $\gamma$ 
specifies the resolution or ``importance'' of the pivot).
 At the lower part of the table the performance of buy-and-hold (BAH) is specified. 
It is evident that both methods suffer from greater values of $\gamma$ (5\% and 10\%) and very small values of $\gamma$(0.1\%). Specifically, 
for these $\gamma$ values the mean returns are lower than the corresponding return of BAH, and the 
Sharpe ratios do not exceed by much the 
Sharpe ratio of BAH.
With lower values of $\gamma$ (1\% and 2\%) ANN still fails to beat the BAH mean return, 
while SVR is better than the BAH in average, but this advantage is not statistically significant according to 
Wilcoxon signed rank test whose results are summarized in Table~\ref{tab:pValuesSvrVsANN} in the appendix.

With impact pivot points both ANN and SVR achieved good Sharpe ratios that are better than the BAH Sharpe ratio. 
This advantage is statistically significant.
In addition, SVR is superior to ANN in terms of both mean return and the mean Sharpe ratio.
When considering the MDD measure, we observe that both SVR and ANN achieved better (smaller) MDD than the BAH.
The best performing $\gamma$ in terms of Sharpe ratio is $\gamma=2\%$ (for both types of pivot types).
This emphasizes that the proposed models are more useful for shorter term prediction of smaller sized fluctuations.

\noindent
\textbf{Remark:} The $p$-values corresponding of the Wilcoxon signed-rank test of this experiment appear 
in Table \ref{tab:pValuesSvrVsANN} in the Appendix.

	\begin{table}[htbp]
	\footnotesize
  \centering
    \begin{tabular}{rrrrr}
    \addlinespace
    \toprule
     & Mean Return
     \footnote{Just to get impression what is the yearly return corresponding to these 60-days returns: 1.81\% in 60 days is 12.08\% of annual return, 1.06\% -- 1.3\% of annual return, 1.44\% is 4.6\% annualy.}
      & Success Rate & Mean MDD & Mean ASR\\
  
    \midrule
 	  \multicolumn{5}{l} {Turning points of impact $\gamma$} \\
    \midrule
    $\gamma$=0.1\%   &       &       &       &  \\
    SVR   & $1.28\%\pm0.30\%$ & $68.02\%\pm1.66\%$ & \textbf{$2.17\%\pm0.20\%$} & $0.94\pm0.13$ \\
    ANN   & $1.10\%\pm0.25\%$ & $61.08\%\pm0.97\%$ & \textbf{$1.75\%\pm0.12\%$} & $0.91\pm0.12$ \\
    $\gamma$=1\%   &       &       &       &  \\
    SVR   & $1.81\%\pm0.28\%$(*) & $69.89\%\pm1.83\%$ & \textbf{1.92\%$\pm$0.18\%} & \textbf{1.34$\pm$0.12}(*) \\
    ANN   & $1.06\%\pm0.22\%$ & $62.76\%\pm1.27\%$ & \textbf{1.98\%$\pm$13.00\%} & $1.14\pm0.11$ \\
    $\gamma$=2\%   &       &       &       &  \\
    SVR   & $1.77\%\pm0.22\%$(*) & $73.82\%\pm1.87\%$ & \textbf{1.73\%$\pm$0.18\%} & \textbf{1.67$\pm$0.13}(*) \\
    ANN   & $1.02\%\pm0.18\%$ & $66.17\%\pm1.39\%$ & \textbf{1.91\%$\pm$0.15\%} & \textbf{1.13$\pm$0.10} \\
    $\gamma$=5\%   &       &       &       &  \\
    SVR   & $0.59\%\pm0.20\%$ & $64.58\%\pm2.31\%$ & \textbf{1.37\%$\pm$0.18\%} & 1.26$\pm$0.13 \\
    ANN   & $0.65\%\pm0.13\%$ & $63.47\%\pm1.79\%$ & \textbf{1.27\%$\pm$0.12\%} & $1.04\pm0.11$ \\
    $\gamma$=10\%  &       &       &       &  \\
    SVR   & $0.54\%\pm0.16\%$ & $76.10\%\pm2.31\%$ & \textbf{0.63\%$\pm$0.13\%} & $1.79\pm0.12$ \\
    ANN   & $0.37\%\pm0.09\%$ & $61.27\%\pm2.04\%$ & \textbf{0.57\%$\pm$0.07\%} & $0.81\pm0.11$ \\
		\midrule
 	  \multicolumn{5}{l} {Turning points of momentum $\gamma$ with respect to lookahead window $w=6$ } \\
    \midrule
    $\gamma=$0.1\%   &       &       &       &  \\
    SVR   & $1.11\%\pm0.28\%$ & $67.11\%\pm1.65\%$ & \textbf{$2.22\%\pm0.20\%$} & $0.82\pm0.12$ \\
    ANN   & $0.98\%\pm0.24\%$ & $61.21\%\pm0.98\%$ & \textbf{$1.71\%\pm0.12\%$} & $0.72\pm0.11$ \\
 	  $\gamma=$1\%   &       &       &       &  \\
    SVR   & $1.46\%\pm0.28\%$(*) & $67.81\%\pm2.01\%$ & \textbf{1.65\%$\pm$0.18\%} & $1.26\pm0.13$(*) \\
    ANN   & $1.07\%\pm0.22\%$ & $62.26\%\pm1.27\%$ & \textbf{1.90\%$\pm$13.00\%} & $1.03\pm0.11$ \\
    $\gamma=$2\%   &       &       &       &  \\
    SVR   & $1.43\%\pm0.26\%$(*) & $69.97\pm2.26\%$ & \textbf{1.64\%$\pm$0.17\%} & \textbf{1.4$\pm$0.13}(*) \\
    ANN   & $0.97\%\pm0.18\%$ & $65.82\%\pm1.39\%$ & \textbf{1.67\%$\pm$0.15\%} & $1.09\pm0.10$ \\
    $\gamma=$5\%   &       &       &       &  \\
    SVR   & $0.82\%\pm0.15\%$ & $75.47\%\pm2.20\%$ & \textbf{0.65\%$\pm$0.11\%} & $1.63\pm0.13$ \\
    ANN   & $0.48\%\pm0.13\%$ & $58.51\%\pm1.79\%$ & \textbf{1.09\%$\pm$0.12\%} & $0.51\pm0.11$ \\
    $\gamma=$10\%  &       &       &       &  \\
    SVR   & $0.23\%\pm0.06\%$ & $77.59\%\pm2.35\%$ & \textbf{0.23\%$\pm$0.07\%} & $1.44\pm0.09$ \\
    ANN   & $0.37\%\pm0.09\%$ & $55.55\%\pm2.04\%$ & \textbf{0.61\%$\pm$0.07\%} & $0.69\pm0.11$ \\
   \bottomrule
 \multicolumn{5}{l} {BAH mean return:$1.44\pm0.44\%$} \\
    \multicolumn{5}{l} {BAH mean MDD:$8.2\pm0.41\%$}\\
    \multicolumn{5}{l} {BAH mean ASR:$1.01\pm0.15\%$}\\
   \bottomrule
    \end{tabular}
   \caption{Performance comparison between the SVR and ANN models for turning points with impact and momentum characteristics. 
   Whenever a result of one of the methods is better at 5\% level, according to Wilcoxon signed-rank test, we mark it with (*). 
    Boldface numbers mark results that exceed the corresponding BAH metric at 5\% level.}
  \label{tab:TPPvsSVRresults}
\end{table}

\subsubsection {Experiment 2: SVR Backward window size}
\label{sec:BackwardWindowSizeExperiment}
We tested the SVR performance for short (4), medium (8) and large (50) backward windows lengths. Table \ref{tab:InputLengthComparison} presents the results of these tests for three types of turning points (2\%-momentum, 2\%-impact and pivots of degree 10). 
Table \ref{tab:BackwardComparisonRanks} in the appendix shows a comparison of the 
SVR return, success rate, maximum drawdown and Sharpe ratio, for the three backward window length, 
with the corresponding metrics of BAH. that table also includes 
the Friedman ranks statistics of these tests. 
From this comparative analysis we can conclude that the small and medium backward window lengths 
allow SVR to perform  significantly better 
than BAH in terms of Sharpe ratio. However, according to the Friedman ranks analysis, 
small and medium ASR belong to the  same performance group.
this means that no statistically significant difference between them was detected in the post-hoc test.

\begin{table}[htbp]
	\footnotesize
  \centering
    \begin{tabular}{rrrrr}
    \addlinespace
    \toprule
    Backward window & Mean return & Success Rate & Mean MDD & Mean ASR\\
    \midrule
    \multicolumn{4}{l} {Turning points with impact $\gamma=2\%$} \\
    4     & $1.95\%\pm0.28\%$ & $65.84\%\pm1.74\%$ & \textbf{2.09\%$\pm$0.22\%} & \textbf{1.36$\pm$0.12} \\
    8     & $1.77\%\pm0.22\%$ & $73.82\%\pm1.87\%$ & \textbf{1.73\%$\pm$0.18\%} & \textbf{1.67$\pm$0.12} \\
    50    & $1.36\%\pm0.22\%$ & $70.44\%\pm2.27\%$ & \textbf{0.95\%$\pm$0.14\%} & $1.36\pm0.13$ \\
          &       &       &       &  \\
    \midrule
    \multicolumn{4}{l} {Turning points with momentum $\gamma=2\%$, lookahead window 6 days} \\
    4     & $1.59\%\pm0.27\%$ & $66.41\%\pm1.71\%$ & \textbf{1.99\%$\pm$0.17\%} & \textbf{1.21$\pm$0.13} \\
    8     & $1.43\%\pm0.26\%$ & $69.97\pm2.26\%$ & \textbf{1.64\%$\pm$0.17\%} & \textbf{1.4$\pm$0.13} \\
    50    & $1.39\%\pm0.22\%$ & $69.90\%\pm2.32\%$ & \textbf{0.93\%$\pm$0.14\%} & $1.77\pm0.13$ \\
    &       &       &       &  \\
    \midrule
    \multicolumn{4}{l} {Pivot turning points of degree 10} \\
    4     & $1.21\pm0.28\%$ & $64.76\pm1.92\%$ & \textbf{0.11$\pm$0.00} & \textbf{2.32$\pm$0.22\%} \\
    8     & $1.23\pm0.20\%$ & $69.97\pm2.26\%$ & \textbf{0.26$\pm$0.12} & \textbf{1.33$\pm$0.15\%} \\
    50    & $0.35\pm0.31\%$ & $53.87\pm2.59\%$ & \textbf{0.09$\pm$0.00} & 1.50$\pm$0.25\% \\

    &       &       &       &  \\
\bottomrule
    \multicolumn{5}{l} {BAH mean return:$1.44\pm0.44\%$} \\
    \multicolumn{5}{l} {BAH mean MDD:$8.2\pm0.41\%$}\\
    \multicolumn{5}{l} {BAH mean ASR:$1.01\pm0.15\%$}\\
   \bottomrule
    \end{tabular}
   \caption{Performance dependence on backward window length. Boldface numbers are significantly better than the corresponding 
   BAH metrics. The $p$-values of the pairwise tests with BAH and Friedman ranks are summarized in Table \ref{tab:BackwardComparisonRanks}.}
  \label{tab:InputLengthComparison}
\end{table}

\subsubsection {Experiment 3: SVR training segment length}
\label{sec:TrainingSegmentLengthExperiment}
The goal of this experiment was to see how performance is influenced by the training segment length. 
To this end,  we considered three lengths: short (0.5 year), medium (1 and 2 years) and long (5 years).
The results are presented in the Table \ref{tab:TrainingSetLengthComparison}.
Based on the statistical analysis of these results (summarized in Table \ref{tab:TrainsetLenComparison} in the appendix), 
we conclude that the algorithm exhibits similar performance for these four training segment lengths,
and this holds for the three types of pivot points (pivot degree, momentum and impact).
While the average Sharpe ratios and returns are higher for longer training segments, 
the Friedman rank analysis cannot designate these differences statistically significant.

	\begin{table}[htbp]
	\footnotesize
  \centering
    \begin{tabular}{rrrrr}
    \addlinespace
    \toprule
    Training set length & Mean return & Success Rate & Mean MDD & Mean ASR\\
    \midrule
    \multicolumn{4}{l} {Impact turning points} \\
    0.5 year & $1.77\%\pm0.25\%$ & 66.84\%$\pm$2.19\% & {\bf 1.86\%$\pm$0.18\%} & {\bf 1.36$\pm$0.13} \\
    1 year & $1.47\%\pm0.22\%$ & 70.95\%$\pm$2.05\% & {\bf 1.79\%$\pm$0.18\%} & {\bf 1.38$\pm$0.12} \\
    2 years & $1.77\%\pm0.22\%$ & 73.82\%$\pm$1.87\% & {\bf 1.73\%$\pm$0.18\%} & {\bf 1.67$\pm$0.12} \\
    5 years & $2.12\%\pm0.26\%$ & 73.92\%$\pm$1.97\% & {\bf 1.90\%$\pm$0.19\%} & {\bf 1.67$\pm$0.12} \\

          &       &       &       &  \\
    \midrule
    \multicolumn{4}{l} {Momentum turning points} \\

    0.5 year & $1.07\%\pm0.27\%$ & $68.00\%\pm1.90\%$ & \textbf{1.66\%$\pm$0.18\%} & \textbf{1.23$\pm$0.13} \\
    1 year & $1.42\%\pm0.27\%$ & $68.52\%\pm1.80\%$ & \textbf{1.66\%$\pm$0.17\%} & \textbf{1.25$\pm$0.13} \\
    2 years   & $1.43\%\pm0.26\%$ & $69.97\pm2.26\%$ & \textbf{1.64\%$\pm$0.17\%} & \textbf{1.4$\pm$0.13} \\
    5 years & $1.36\%\pm0.28\%$ & $65.44\%\pm2.13\%$ & \textbf{1.63\%$\pm$0.19\%} & 1.15$\pm$0.13 \\

    &       &       &       &  \\
    \midrule
    \multicolumn{4}{l} {Pivot degree turning points} \\

    0.5 year & $0.63\%\pm0.20\%$ & $63.63\%\pm2.38\%$ & 1.63\%$\pm$0.20\% & 0.96$\pm$0.12 \\
    1 year & $1.07\%\pm0.24\%$ & $65.97\%\pm2.42\%$ & 1.72\%$\pm$0.20\% & 1.23$\pm$0.13 \\
    2 years & $1.23\%\pm0.20\%$ & $69.97\%\pm2.26\%$ & 1.33\%$\pm$0.15\% & {\bf 1.45$\pm$0.12} \\
    5 years & $1.08\%\pm0.23\%$ & $70.18\%\pm2.32\%$ & 1.44\%$\pm$0.18\% & 1.35$\pm$0.14 \\

    &       &       &       &  \\
\bottomrule
    \multicolumn{5}{l} {BAH mean return:$1.44\pm0.44\%$} \\
    \multicolumn{5}{l} {BAH mean MDD:$8.2\pm0.41\%$}\\
    \multicolumn{5}{l} {BAH mean ASR:$1.01\pm0.15\%$}\\
   \bottomrule
    \end{tabular}
   \caption{Performance dependence on training segment length}
  \label{tab:TrainingSetLengthComparison}
\end{table}

\subsubsection {Experiment 4: reproduction of  the Li et al. results in \cite{LiDL09a}}
In this experiment the goal was to reproduce the results presented in \cite{LiDL09a} for the original ANN model
with respect to the DJIA and TESCO price sequences.
Due to the use of randomization in the backpropagation training algorithm 
(used for random assignment of initial weights), we ran each experiment 
20 times in order to receive a robust estimation of the result. The results are presented In 
Table~\ref{tab:Table1SeveralRunsOfTheAlgorithmOnTheDJIADataset} (containing sub-tables (a) for DJIA, and (b)
for TESCO). All these experiments consider the same training/validation/test partition.\footnote{For the DJIA sequence, the training period is 7/9/2005-10/11/2005, the validation period is 11/11/2005-29/1/2006, 
and the test period is 30/1/2006-25/4/2006; for TESCO sequence the training period is 
18/11/2004-31/-5/2006, the validation period is 1/06/2006-26/07/3006, and the test period is 27/7/2006-9/10/2006.}

As expected, due to this randomized ANN training, the 20 the returns obtained (for each dataset) 
vary considerably.
The results reported in \cite{LiDL09a} are quite close to the best return among the 20.
Specifically, Li et al. report on 6.21 return for DJIA (vs. our 5.97), and 11.63 return for TESCO (vs. our 10.94).

The average returns we obtained in both experiments are, unfortunately, not as optimistic as reported in \cite{LiDL09a}
and, in particular, in our experiments the ANN model could not outperform BAH. 
Using the SVR model over these test segments we obtained a return of 2.31\% for DJIA and 7.83\% for TESCO. 
These results are robust since no random selections are made in our SVR training and prediction.

\begin{table}
\scriptsize
\label{tab:Table1SeveralRunsOfTheAlgorithmOnTheDJIADataset}
\centering
\subfloat[DJIA dataset]{
       \begin{tabular}{|r|r|}
       \hline
			 Return(\%) & Success Rate \\
			 \hline
      1.05  &    66.67\% \\
		   3.85  &  100.00\% \\
			 3.05  &   100.00\% \\
			 2.91  &  100.00\% \\
			 4.68  &   75.00\% \\
			 3.26  &    75.00\% \\
			 5.57  &   100.00\% \\
			 3.82  &   66.67\% \\
			 2.56  &   66.67\% \\
			 4.50  &  100.00\% \\
			 \textbf{5.97}  & \textbf{100.00\%} \\
			 2.80  &  100.00\% \\
			 2.35  &  100.00\% \\
			 5.42  &  100.00\% \\
			 1.91  &  66.67\% \\
			 3.90  &  100.00\% \\
			 2.96  &  75.00\% \\
			 0.81  &   66.67\% \\
			 4.14  &  100.00\% \\
			 3.85  &  75.00\% \\
		   \hline
			 \multicolumn{2}{|c|}{Average TPP return:$1.39\%\pm0.31\%$} \\
			 \multicolumn{2}{|c|}{BAH:  $3.91\%$} \\
			 \hline
       \end{tabular}
}
\qquad\qquad
\subfloat[TESCO dataset]{        
       \begin{tabular}{|r|r|}
              \hline
			 Return(\%) & Success Rate \\
			 \hline
       0.00 &      NaN\\
				6.86 &	100.00\%\\
				8.86 &	100.00\%\\
				0.00 &       NaN\\
				7.96 &	100.00\%\\
				3.53 &	100.00\%\\
				9.42 &	100.00\%\\
				0.00 &	        NaN\\
				0.00 &	        NaN\\
				3.46 &	100.00\%\\
				7.62 &	100.00\%\\
				0.00 &	        NaN\\
				0.00 &	        NaN\\
				0.00 &	        NaN\\
				7.06 &	100.00\%\\
				1.59 &	100.00\%\\
				5.82 &	75.00\%\\
				\textbf{10.94} &	\textbf{100.00\%}\\
				7.89 &	100.00\%\\
				1.73 &	100.00\%\\
		   \hline
			 \multicolumn{2}{|c|}{Average TPP return:$4.14\%\pm0.85\%$} \\
			 \multicolumn{2}{|c|}{BAH: $6.37\%$} \\
			 \hline
       \end{tabular}
}
\caption{Attempting to replicate the \cite{LiDL09a} results: 20 iterations of the ANN model on the DJIA and TESCO datasets }
\end{table}

\subsubsection {Experiment 5: ANN ensemble size}
\label{sec:EnsembleResults}
The goals of this experiment were to test how the ANN performance is influenced by the ensemble size.
In particular, we were also interested to see if a single network can perform as well as an ensemble.
In the first experiment we ran the ANN model with a single network, with varying number of neurons in the hidden layer.
In the second experiment we ran 
the ANN model with different ensemble sizes. In this case, the number of neurons 
in the hidden layer of each ensemble member was taken from the interval $[m,m+n]$ where $m$ is the backward window length 
and $n$ is the ensemble size.\footnote{This choice was not explicitly mentioned in 
\cite{LiDL09a} but was obtained from Li et al. by personal communication.}

Table \ref{tab:EnsembleResults} summarizes the performance of these settings.
These results confirm that an ensemble, containing multiple networks outperforms a model with a single network. 
As for the ensemble size, in a direct comparison, no statistically significant difference between ensembles of sizes 2, 5, 10 or 50 members was observed. Nevertheless, the ensemble of size 10 excelled, compared to the other three ensembles, in that it was the only one which
outperformed BAH in terms of Sharpe ratio in a statistically significant manner (ensembles with 5,10 and 50 members obtained better Sharpe ratio than BAH).

\begin{table}[htbp]
  \centering
    \begin{tabular}{rrrrr}
    \addlinespace
    \toprule
    	 & Mean return & Success Rate & Mean MDD & Mean ASR\\
    \midrule      
    \multicolumn{5}{l} {Single network} \\
    
    \midrule      
    \multicolumn{4}{l} {Neurons number}\\

       2     & 0.09\%$\pm$0.07\% & 64.58\%$\pm$6.29\% & 0.15\%$\pm$0.05\% & 1.21$\pm$0.28 \\
    5     & 0.30\%$\pm$0.12\% & 58.90\%$\pm$5.11\% & 0.43\%$\pm$0.12\% & 1.00$\pm$0.24 \\
    10    & 0.68\%$\pm$0.31\% & 64.46\%$\pm$3.42\% & 1.00\%$\pm$0.15\% & 0.83$\pm$0.23 \\
    50    & 0.88\%$\pm$0.42\% & 53.13\%$\pm$1.50\% & 1.59\%$\pm$0.21\% & 0.59$\pm$0.23 \\
    \midrule
    \multicolumn{5}{l} {Ensemble of networks} \\
    \midrule      
    \multicolumn{4}{l} {Members}\\
      2     & 0.57\%$\pm$0.20\% & 62.18\%$\pm$4.02\% & 0.58\%$\pm$0.10\% & 1.15$\pm$0.22 \\
    5     & 1.33\%$\pm$0.27\% & 63.60\%$\pm$3.56\% & 0.95\%$\pm$0.14\% & 1.27$\pm$0.23 \\
    10    & 1.32\%$\pm$0.33\% & 69.35\%$\pm$3.01\% & 1.22\%$\pm$0.19\% & \textbf{1.38$\pm$0.23} \\
    50    & 1.35\%$\pm$0.37\% & 63.68\%$\pm$2.15\% & 1.25\%$\pm$0.14\% & 1.11$\pm$0.22 \\

\bottomrule
    \multicolumn{5}{l} {BAH mean return:$1.18\pm0.87\%$} \\
    \multicolumn{5}{l} {BAH mean MDD:   $6.59\pm0.55\%$} \\
    \multicolumn{5}{l} {BAH mean ASR:   $0.78\pm0.3$} \\
   \bottomrule
    \end{tabular}
   \caption{Performance of the ANN model with a single network constructed with various hidden layer sizes;
   performance of the ANN model with various ensemble sizes}
  \label{tab:EnsembleResults}
\end{table}

\subsubsection {Some caveats regarding performance calculation}

Here we discuss caveats that should be accounted for in examining the above results.
Our experimental protocol examined the trading performance during numerous test segments each of which fixed in length (60 days).
Each trade is completed only when both entry and exit signals are generated.
The first problematic situation occurred in instances where
an entry signal was generated during the test segment but no matching exit signal was obtained during the same test segment.
We could treat such instances in various ways (e.g., discard the trade, end it prematurely, or follow it till it end in another 
segment).
For simplicity of implementation, in our statistics we ignored such trades.
after observing that the results with or without them are very close.


The second problematic situation is when no signals were generated at all during a test segment. 
In such instances the return of the strategy and the maximum drawdown are zero, while the Sharpe ratio is 
not well defined, because the standard deviation of the cumulative 
return (i.e., risk) is zero as well. Hence, we omitted such instances when calculating and averaging Sharpe ratios.
The same consideration and treatment was applied to the success rate metric. Here, when no trades are performed during a test segment,
the success rate is 100\% but is vacuous. Therefore, to be fair, we discarded these periods as well when averaging the
success rate.
An alternative treatment in the case of the Sharpe ratio could be to 
define the Sharpe ratio of such periods as zero and include it in the statistics.
Similarly, we could define the success rate of empty segments as zero as well. However,
this treatment does not make much sense.\footnote{If we do count the Sharpe ratio of such test periods as zero, the resulting average Sharpe ratio of SVR and ANN do not perform significantly better than the average BAH Sharpe.}

To quantify the extent of test periods with undefined Sharpe ratios, consider two settings of SVR prediction
defined with impact turning points of resolutions $\gamma = 0.01$ and $\gamma = 0.02$ (see Table~\ref{tab:TPPvsSVRresults}). 
For 0.01-impact turning points there are 23 test segments without trades (8\% of all segments).
The average BAH return and Sharpe ratio over these segments only are 0.9\% and 1.19, respectively. 
If we discard these segments when calculating the SVR mean Sharpe ratio we obtain 1.34. However, if we account these 
undefined Sharpe ratio as zeros, the result is 0.8. The BAH Sharpe ratio for all test segments is 1.01. 
When considering the setting with 0.02-impact turning points, there are 67 segments (22\%) without trades (we have more such segments 
in this setting because turning points occur less frequently so there are less trades).
The average BAH return and Sharpe ratio for these empty segments are 1.2\%, and 1.12, respectively. 
When we substitute zero for these undefined Sharpe ratios the resulting mean Sharpe ratio for SVR 
in all periods is 1.02, as opposed to 1.67 when ignoring these empty segments. 

For ANN prediction with impact turning points the analogous results are as follows.
 For $\gamma = 0.01$, there are (on average) 26.4 test segments without any trades.
  The average BAH return and Sharpe ratio over these segments are 2.68\% and  1.48, respectively. 
  If we discard these segments, the Sharpe ratio obtained is 1.14. However, if we consider these Sharpe ratios as zeros, 
  the result is 0.6. When considering the setting with 0.02 impact turning points, there are 51.4 segments without trades on average.
  For these segments the average BAH return is 2.32\%, the average Sharpe ratio is 1.36. When takeing these Sharpe ratios as zeros, 
  the resulting ratio is 0.67. When ignoring these segments the average Sharpe ration is 1.13.

\section {Concluding remarks}

Drawing on, and extending useful ideas of Li et al. \cite{LiDL09a}, we proposed and studied
a prediction model for turning points that relies on support vector regression (SVR).
Extensive empirical examination of the proposed model showed that it outperforms the Li et al. neural network model for the
same prediction problem. This advantage is statistically significant.
While our SVR model achieves higher average return than the buy-and-hold benchmark, the difference between 
the two is not statistically significant. However, the SVR model suffers from
significantly lower drawdowns and exhibits higher risk adjust return (Sharpe ratio) than the buy-and-hold.
This significantly lesser risk may give rise to substantially larger return as well using leverage.

Our studies included a complete reproduction and a thorough analysis of the ANN prediction model of Li et al. \cite{LiDL09a}.
This model is considered the state-of-the-art in turning point prediction.  
Our tests included an empirical valuation of that model over multiple periods 
(as opposed to a single period in the original paper).\footnote{More precisely, the Li et al. empirical evaluation 
was confined to two price sequences (DJIA and TESCO) and for each sequence a single test period was considered.}
Our conclusions differ than those of  Li et al. in regards to the performance of their model, 
and in particular, are less optimistic than their conclusions.
Nevertheless, their ANN model contains interesting and useful ideas that have been utilized here and laid the foundations of the present work.

Our work can be extended and modified in various interesting ways. 
First, it would be interesting to see if better representation can be constructed that utilizes
other price sequences and economical indicators. Such inter-market models are considered be more powerful than autoregressive models.
It would also be very interesting to examine lower time-frames such as hourly prices. While intraday data is considered to 
be more noisy, it contain more fluctuations that could be identified by our model. 
Finally, it would be interesting to include rejection mechanisms in the spirit of Chow 
(see, e.g., \cite{Chow,ElYanivWiener}
that can increase prediction accuracy by avoiding prediction in cases of uncertainty.

\bibliographystyle{elsarticle-num}
\bibliography{PaperBib}

\clearpage
\appendix
\label{app:DescriptionOfANN}

\section{An algorithm for extracting alternating pivots}
\label{sec:TPExtractionDetails}

We are given a time series and would like to extract an alternating sequence of peaks and troughs that satisfy a given property. Throughout the 
description of the algorithm we only consider pivots that satisfy the given property.
The first phase of this procedure extracts an alternating sequence of peaks/troughs. The second phase improves the outcome by considering
better (higher or lower) alternatives for each pivot.
\paragraph {Extract initial alternating sequence of peaks and troughs} 
	\begin{enumerate}
			\item Identify the first pivot in the sequence and assume w.l.o.g. that it is a trough $T_1$.
			\item Find the subsequent peak $P_1$.
			\item Discard all troughs appearing between the trough $T_1$ and the peak $P_1$. Denote by $D$
			   the set of all discarded peaks and troughs.
			\item Let $T_2$ be the subsequent trough. Discard all peaks  between $P_1$ and $T_2$.
			\item Repeat these steps steps until there are no more peaks and troughs.
		\end{enumerate}

\paragraph { Peaks/Troughs improvement}
Consider the alternating sequence of peaks and troughs computing as described above.
For each pair of subsequent troughs $T_i$ and $T_j$ there exists a unique peak $P_k$ appearing between the $T_i$ and $T_j$ (due to alternation). 
Replace $P_k$ by $P_{k'}$ if $P_k < P_{k'}$ (i.e. $P_{k'}$ is a higher peak) and $P_{k'} \in D$ (i.e., it was discarded in step (3) of the 
first phase). Apply the analogous procedure to improve troughs.

\clearpage
\section{Additional experimental details}
Here we provide complimentary details for the experiments described in 
sections \ref{sec:SVRvsANN}, \ref{sec:BackwardWindowSizeExperiment}, and 
\ref{sec:TrainingSegmentLengthExperiment}. 

\paragraph {Experiment 1: SVR vs. ANN vs. BAH.}
In Experiment 1(Section~\ref{sec:SVRvsANN}) we compared the performance of the SVR model to the performance of the ANN model. In addition, we compared each of these algorithms to the BAH benchmark.  In Table~\ref{tab:pValuesSvrVsANN} the $p$-values for the  Wilcoxon signed rank test are presented 
for each of the three comparisons (ANN vs. SVR, BAH vs. ANN, and BAH vs. SVR), 
and for each of the performance metrics (Mean Return, Mean MDD, Mean ASR). Whenever the $p$-value 
obtained in the test is less than 0.05, we can conclude that the performance of two compared 
algorithms is significantly different (such $p$-values appear in \textbf{boldface} font), otherwise we conclude that the performance is statistically indistinguishable at 95\% confidence level. 
\begin{table}[htbp]
  \centering
    \begin{tabular}{rrrr}
    \addlinespace
    \toprule
    pValues & Return & MDD   & ASR \\
    \midrule
    ANN vs SVR &       &       &  \\
    0.1\% & 0.6684 & \textbf{0.03} & 0.4798 \\
    1\%   & \textbf{0.0101} & \textbf{0.019} & \textbf{0.035} \\
    2\%   & \textbf{0.029} & \textbf{0.013} & \textbf{0.044} \\
    5\%   & 0.522 & 0.162 & \textbf{0.0464} \\
    10\%  & \textbf{0.0192} & 0.21 & 0.51 \\
          &       &       &  \\
    BAH vs ANN &       &       &  \\
    0.1\% & 0.595 & \textbf{2.27E-15} & 0.5073 \\
    1\%   & 0.947783 & \textbf{4.25E-16} & 0.055 \\
    2\%   & 0.865876 & \textbf{8.00E-17} & \textbf{0.0135} \\
    5\%   & 0.909438 & \textbf{7.55E-17} & 0.0954 \\
    10\%  & 0.722563 & \textbf{3.17E-16} & 0.8875 \\
    			&       &       &  \\
    BAH vs SVR &       &       &  \\
    0.1\% & 0.375 & 2.09E-9 & 0.8899 \\
    1\%   & 0.564861 & \textbf{3.30E-17} & \textbf{0.0028} \\
    2\%   & 0.7482 & \textbf{5.31E-17} & \textbf{2.20E-05} \\
    5\%   & 0.501418 & \textbf{7.98E-17} & 0.0918 \\
    10\%  & {\bf 0.0277} & \textbf{4.98E-21} & 0.265 \\
          &       &       &  \\
    \bottomrule
    \end{tabular}
    \caption{Experiment 1: $p$-values of the Wilcoxon signed-rank test comparing pairwise performance 
    of ANN, SVR and the BAH benchmark.}
  \label{tab:pValuesSvrVsANN}
\end{table}

\paragraph {Experiment 2: SVR Backward window size.}
In Experiment 2(Section~\ref{sec:BackwardWindowSizeExperiment}) we examined the performance of the SVR 
prediction model as a function of the backward window length. This examination was repeated for different pivot point types. 
Statistical analysis was performed separately for each pivot point type. 
Two statistical tests were conducted: the Wilcoxon signed-rank test to compare the performance of SVR and BAH 
(each one of the performance metrics was compared to the corresponding BAH metric), 
and the Friedman test to compare the SVR performance for each pivot point type. 
The Friedman test has two stages. In the first stage, the null hypothesis is that 
``all algorithms have the same performance and the results differ due to a chance.'' If the null hypothesis is not rejected 
at 95\% confidence level, we conclude that all the algorithms in the group are statistically indistinguishable. Otherwise, if the hypothesis is rejected, a post-hoc test is performed that divides the group into two or more subgroups such that in each subgroup the performance is indistinguishable at 95\% level. In case that such a subgroup exist, we denote the best subgroup members with boldface 
in Table~\ref{tab:BackwardComparisonRanks}.

\begin{table}
		\begin{tabular} {lrrrcrrr}
		 \toprule
      & \multicolumn{ 3}{c}{Wilcoxon Signed-Rank Test}& \phantom{} & \multicolumn{ 3}{c}{Friedman ranks} \\
    	\cmidrule {2-4}\cmidrule {6-8} \\
      Backward window & Return & MDD & ASR && Return & MDD & ASR \\
    	    \midrule
    Impact &       &       &       &       &       &  \\
    \midrule
    {\bf 4} & 0.5348 & {\bf 2.15E-48} & {\bf 0.0164} && {\bf 2.09} & {\bf 2.2} & {\bf 2.05} \\
    {\bf 8} & 0.8814 & {\bf 9.65E-43} & {\bf 0.0001} && {\bf 2.04} & {\bf 2.07} & {\bf 2.09} \\
    {\bf 50} & 0.3776 & {\bf 3.67E-38} & 0.128 && 1.87  & 1.73  & 1.86 \\
          &       &       &       &       &       &  \\
    Momentum &       &       &       &       &       &  \\
    \midrule
    {\bf 4} & 0.469 & {\bf 2.12E-40} & {\bf 0.02} && 2.09  & 2.27  & 2.14 \\
    {\bf 8} & 0.0855 & {\bf 2.13E-34} & {\bf 0.0118} && 1.95  & 2.04  & 1.98 \\
    {\bf 50} & 0.0723 & {\bf 1.12E-22} & 0.1717 && 1.96  & {\bf 1.69} & 1.88 \\
          &       &       &       &       &       & \\
    Pivot &       &       &       &       &       & \\
    \midrule
    {\bf 4} & 0.3921 & {\bf 7.84E-39} & {\bf 0.0289} && 2.05  & 2.24  & 2.09 \\
    {\bf 8} & 0.0672 & {\bf 5.73E-31} & {\bf 0.0046} && 2.02  & 2.03  & 2.01 \\
    {\bf 50} & 0.0513 & {\bf 2.30E-21} & 0.6463 && 1.93  & {\bf 1.74} & 1.9 \\
    \bottomrule

		 \end{tabular}

		 \caption{Experiment 2: Performance dependence on backward window length. $p$-values of the BAH comparison, and the
		 Friedman rank test. The values representing statistical significance $p<0.05$ for the Wilcoxon and Friedman post-hoc tests are marked with \textbf{boldface}.}
		\label{tab:BackwardComparisonRanks}		 
\end{table}

\clearpage
\paragraph {Experiment 3: SVR training segment length}
Experiment 3(Section~\ref{sec:TrainingSegmentLengthExperiment}) results presentation is similar to presentation of the results for Experiment 2 in Table~\ref{tab:BackwardComparisonRanks}. The results of Wilcoxon test and Friedman test are summarized in Table~\ref{tab:TrainsetLenComparison}.
\begin{table}[!h]
		\begin{tabular} {lrrrcrrr}
		 \toprule
      & \multicolumn{ 3}{c}{Wilcoxon Signed-Rank Test}& \phantom{} & \multicolumn{ 3}{c}{Friedman ranks} \\
    	\cmidrule {2-4}\cmidrule {6-8} \\
      Train set length & Return & MDD & ASR && Return & MDD & ASR \\
    	    \midrule

    Impact &       &       &       &       &       &       &  \\
    \midrule
    0.5 year & 0.9101 & {\bf 6.33E-44} & {\bf 0.0077} & {\bf } & 2.62  & 2.51  & 2.53 \\
    1 year & 0.7206 & {\bf 1.37E-46} & {\bf 0.0045} & {\bf } & 2.46  & 2.49  & 2.43 \\
    2 years & 0.8814 & {\bf 9.65E-43} & {\bf 0.0001} & {\bf } & 2.36  & 2.54  & 2.44 \\
    5 years & 0.2713 & {\bf 8.13E-44} & {\bf 0.0001} & {\bf } & 2.56  & 2.46  & 2.6 \\
          &       &       &       &       &       &       &  \\
    Momentum   &       &       &       &       &       &       &  \\
    \midrule
    0.5 year & 0.2074 & {\bf 5.02E-33} & {\bf 0.0488} & {\bf } & 2.45  & 2.63  & 2.4 \\
    1 year & 0.1389 & {\bf 2.66E-35} & {\bf 0.004} & {\bf } & 2.52  & 2.42  & 2.55 \\
    2 years & 0.0855 & {\bf 2.13E-34} & {\bf 0.0118} & {\bf } & 2.4   & 2.55  & 2.4 \\
    5 years & 0.5036 & {\bf 2.48E-33} & {\bf 0.003} & {\bf } & 2.63  & 2.4   & 2.64 \\
          &       &       &       &       &       &       &  \\
    Pivot &       &       &       &       &       &       &  \\
    \midrule
    0.5 year & 0.1771 & {\bf 2.94E-29} & {\bf 0.009} & {\bf } & 2.53  & 2.49  & 2.42 \\
    1 year & 0.2106 & {\bf 2.78E-31} & {\bf 0.002} & {\bf } & 2.59  & 2.53  & 2.59 \\
    2 years & 0.0672 & {\bf 5.73E-31} & {\bf 0.0046} & {\bf } & 2.44  & 2.51  & 2.50 \\
    5 years & 0.0529 & {\bf 7.57E-27} & 0.0557 &       & 2.44  & 2.47  & 2.49 \\
          &       &       &       &       &       &       &  \\
          &       &       &       &       &       &       &  \\
    \bottomrule

	 \end{tabular}

		 \caption{Experiment 3: Performance dependence on training segment length. Left part contains p-values for BAH comparison, right part contains the Friedman ranks. Whenever statistical significant difference is achieved, the results are marked with boldface.}
		\label{tab:TrainsetLenComparison}		 
\end{table}

\clearpage
\paragraph {Experiment 5: ANN ensemble size}
Table \ref{tab:chaoticPropertiesDatasets} contains the data segments along with their chaotic properties that were used in the experiment described in Section~\ref{sec:EnsembleResults}.

\begin{table}[htbp]
\label{tab:chaoticPropertiesDatasets}
	\tiny
  \centering
     \begin{tabular}{rrrrrr}
    \addlinespace
    \toprule
    Training start &       Test start &  EmbeddingDimension &       Delay &         LLE & Look-ahead\\
    \midrule
                23-Sep-1958 &      29-Feb-1960 & 7     & 3     & 0.17  & 6 \\
        01-Sep-1960 &      08-Feb-1962 & 6     & 3     & 0.16  & 6 \\
        25-Jul-1961 &      28-Dec-1962 & 7     & 5     & 0.19  & 5 \\
        22-Jul-1963 &      24-Dec-1964 & 6     & 5     & 0.15  & 7 \\
        26-Jul-1963 &      31-Dec-1964 & 6     & 3     & 0.14  & 7 \\
        20-Aug-1964 &      25-Jan-1966 & 5     & 2     & 0.19  & 5 \\
        31-Jan-1966 &      06-Jul-1967 & 6     & 3     & 0.18  & 6 \\
        03-Jul-1967 &      15-Jan-1969 & 5     & 3     & 0.22  & 5 \\
        13-Dec-1967 &      30-Jun-1969 & 6     & 2     & 0.19  & 5 \\
        20-Feb-1970 &      26-Jul-1971 & 7     & 2     & 0.16  & 6 \\
        13-Jan-1971 &      14-Jun-1972 & 5     & 4     & 0.21  & 5 \\
        15-Jan-1971 &      16-Jun-1972 & 5     & 4     & 0.21  & 5 \\
        18-Mar-1971 &      17-Aug-1972 & 6     & 4     & 0.21  & 5 \\
        23-Aug-1972 &      30-Jan-1974 & 6     & 4     & 0.21  & 5 \\
        22-Nov-1972 &      01-May-1974 & 7     & 3     & 0.2   & 5 \\
        12-Apr-1973 &      16-Sep-1974 & 6     & 5     & 0.21  & 5 \\
        16-May-1974 &      17-Oct-1975 & 7     & 2     & 0.18  & 6 \\
        22-Jul-1974 &      22-Dec-1975 & 7     & 2     & 0.19  & 5 \\
        10-Feb-1975 &      14-Jul-1976 & 7     & 3     & 0.18  & 6 \\
        04-Mar-1975 &      04-Aug-1976 & 5     & 3     & 0.17  & 6 \\
        11-Nov-1975 &      15-Apr-1977 & 5     & 2     & 0.18  & 6 \\
        27-Dec-1976 &      31-May-1978 & 5     & 6     & 0.17  & 6 \\
        27-Mar-1978 &      27-Aug-1979 & 5     & 2     & 0.22  & 5 \\
        24-Apr-1978 &      25-Sep-1979 & 5     & 2     & 0.22  & 5 \\
        05-Feb-1979 &      09-Jul-1980 & 5     & 2     & 0.2   & 5 \\
        28-Sep-1979 &      04-Mar-1981 & 7     & 3     & 0.18  & 6 \\
        19-Feb-1982 &      22-Jul-1983 & 6     & 3     & 0.16  & 6 \\
        31-Aug-1982 &      01-Feb-1984 & 6     & 2     & 0.16  & 6 \\
        02-Sep-1983 &      05-Feb-1985 & 5     & 2     & 0.2   & 5 \\
        12-Nov-1984 &      18-Apr-1986 & 6     & 3     & 0.16  & 6 \\
        19-Dec-1986 &      24-May-1988 & 6     & 1     & 0.16  & 6 \\
        20-Jan-1988 &      22-Jun-1989 & 6     & 4     & 0.17  & 6 \\
        27-Feb-1989 &      31-Jul-1990 & 5     & 2     & 0.19  & 5 \\
        22-Dec-1989 &      29-May-1991 & 6     & 3     & 0.19  & 5 \\
        27-Aug-1990 &      29-Jan-1992 & 5     & 1     & 0.16  & 6 \\
        30-Dec-1991 &      02-Jun-1993 & 7     & 3     & 0.18  & 6 \\
        05-Aug-1993 &      09-Jan-1995 & 5     & 4     & 0.17  & 6 \\
        06-Sep-1994 &      07-Feb-1996 & 5     & 3     & 0.15  & 7 \\
        11-May-1995 &      11-Oct-1996 & 6     & 2     & 0.14  & 7 \\
        05-May-1997 &      07-Oct-1998 & 5     & 3     & 0.16  & 6 \\
        13-Oct-1998 &      17-Mar-2000 & 7     & 4     & 0.18  & 6 \\
        28-Jun-2000 &      06-Dec-2001 & 6     & 3     & 0.21  & 5 \\
        01-Apr-2002 &      03-Sep-2003 & 6     & 1     & 0.17  & 6 \\
        23-Aug-2002 &      29-Jan-2004 & 5     & 4     & 0.16  & 6 \\
        29-Apr-2003 &      01-Oct-2004 & 7     & 5     & 0.14  & 7 \\
        22-Apr-2004 &      26-Sep-2005 & 7     & 2     & 0.19  & 5 \\
        24-Jun-2004 &      25-Nov-2005 & 7     & 1     & 0.19  & 5 \\
        25-Oct-2004 &      30-Mar-2006 & 7     & 1     & 0.18  & 6 \\
        25-May-2005 &      27-Oct-2006 & 7     & 2     & 0.17  & 6 \\
        01-Sep-2005 &      08-Feb-2007 & 6     & 2     & 0.18  & 6 \\

    \bottomrule
    \end{tabular}
  \caption{Details of the data segments used for ANN method }
\end{table}

	\begin{table}[htbp]
	\footnotesize
  \centering
    \begin{tabular}{rrrrr}
    \addlinespace
    \toprule
     $\gamma$ & Mean Return & Success Rate & Mean MDD & Mean ASR\\
    	\midrule 
    \multicolumn{5}{l} {Turning points of impact $\gamma$} \\
    1\%   & $1.55\%\pm0.26\%$ & $69.43\%\pm1.97\%$ & {\bf 1.60\%$\pm$0.17\%} & {\bf 1.38$\pm$0.12} \\
    2\%   & $1.54\%\pm0.25\%$ & $74.03\%\pm1.99\%$ & {\bf 1.65\%$\pm$0.18\%} & {\bf 1.62$\pm$0.13} \\
    5\%   & $0.50\%\pm0.21\%$ & $70.06\%\pm2.44\%$ & {\bf 1.26\%$\pm$0.20\%} & 1.25$\pm$0.14 \\
    10\%  & $0.81\%\pm0.18\%$ & $71.22\%\pm2.41\%$ & {\bf 0.61\%$\pm$0.11\%} & 1.61$\pm$0.13 \\
     	\midrule
    \multicolumn{5}{l} {Momentum points of impact $\gamma$} \\
    1\%   & $1.46\%\pm0.28\%$ & $67.81\%\pm2.01\%$ & {\bf 1.65\%$\pm$0.18\%} & {\bf 1.26$\pm$0.13} \\
    2\%   & $1.43\%\pm0.26\%$ & $70.09\%\pm2.21\%$ & {\bf 1.57\%$\pm$0.18\%} & {\bf 1.57$\pm$0.13} \\
    5\%   & $0.82\%\pm0.15\%$ & $75.47\%\pm2.20\%$ & {\bf 0.65\%$\pm$0.11\%} & 1.63$\pm$0.13 \\
    10\%  & $0.23\%\pm0.06\%$ & $77.59\%\pm2.35\%$ & {\bf 0.23\%$\pm$0.07\%} & 1.44$\pm$0.09 \\
  
   \bottomrule
 \multicolumn{5}{l} {BAH mean return:$1.44\pm0.44\%$} \\
    \multicolumn{5}{l} {BAH mean MDD:$8.2\pm0.41\%$}\\
    \multicolumn{5}{l} {BAH mean ASR:$1.01\pm0.15\%$}\\
   \bottomrule
    \end{tabular}
   \caption{Performance comparison between the SVR and ANN models for turning points with impact and momentum characteristics. 
   Whenever a result of one of the methods is better at 5\% level, according to Wilcoxon signed-rank test, we mark it with (*). 
    Boldface numbers mark results that exceed the corresponding BAH metric at 5\% level.}
  \label{tab:TPPvsSVRresults}
\end{table}

\end{document}